\documentclass{article}

\usepackage[english]{babel}
\usepackage{authblk}

\usepackage[table, dvipsnames]{xcolor}
\usepackage{amsmath}
\usepackage{geometry}
\usepackage{graphicx}
\usepackage[colorlinks=true, allcolors=blue]{hyperref}
\usepackage{booktabs}
\usepackage{array}
\usepackage{multirow}

\definecolor{commentgreen}{RGB}{34, 139, 34} 
\usepackage{listings}  
\usepackage{cite}
\usepackage[numbers]{natbib}

\title{Navigating Data Corruption in Machine Learning: Balancing Quality, Quantity, and Imputation Strategies}
\author[1]{Qi Liu*}
\author[2]{Wanjing Ma}
\affil[1,2]{College of Transportation, Tongji University, Shanghai, P.R.China, 201804}
\affil[ ]{\texttt{\{liu\_qi, mawanjing\}@tongji.edu.cn}}

\begin{document}
\maketitle

\begin{abstract}
Data corruption, including missing and noisy entries, is a common challenge in real-world machine learning. This paper examines its impact and mitigation strategies through two experimental setups: supervised NLP tasks (NLP-SL) and deep reinforcement learning for traffic signal control (Signal-RL). This study analyzes how varying corruption levels affect model performance, evaluate imputation strategies, and assess whether expanding datasets can counteract corruption effects. The results indicate that performance degradation follows a diminishing-return pattern, well modeled by an exponential function. Noisy data harms performance more than missing data, especially in sequential tasks like Signal-RL where errors may compound. Imputation helps recover missing data but can introduce noise, with its effectiveness depending on corruption severity and imputation accuracy. This study identifies clear boundaries between when imputation is beneficial versus harmful, and classify tasks as either noise-sensitive or noise-insensitive. Larger datasets reduce corruption effects but offer diminishing gains at high corruption levels. These insights guide the design of robust systems, emphasizing smart data collection, imputation decisions, and preprocessing strategies in noisy environments. The code for reproducing our experiments is available at \url{https://github.com/qiliuchn/data-corruption-study}.
\end{abstract}

\section{Introduction}
Machine learning models rely heavily on high-quality data, yet real-world datasets often suffer from corrupted data—such as missing entries or noisy measurements—which significantly degrades model performance. Data corruption arises from diverse sources, including sensor errors, transmission artifacts, or incomplete data collection. It manifests in two primary forms: missing data (e.g., masked tokens in NLP or undetected vehicles in traffic control) and noisy data (e.g., mislabeled text or perturbed sensor readings). While this challenge is well-documented in classical machine learning (Little and Rubin, 2019 \cite{Little2019}; Emmanuel et al., 2021 \cite{emmanuel2021survey}), its implications for modern paradigms like large language models (LLMs) and deep reinforcement learning (DRL) remain underexplored. 

Different learning paradigms exhibit distinct vulnerabilities to data corruption. For example, LLMs may tolerate certain missing tokens but struggle with semantic noise, whereas DRL policies can compound observational noise over sequential decisions. Prior work has shown that noise in NLP tasks (e.g., mislabeled text) biases model outputs (Brown et al., 2020 \cite{brown2020language}), whereas in DRL, observational noise disrupts policy stability (Pathak et al., 2017  \cite{pathak2017curiosity}). Despite these insights, fundamental questions remain unanswered: How do corruption type and severity impact different learning paradigms? Can imputation effectively recover lost performance, or does it can make things worse by introducing new noise? And when is collecting more data a viable alternative to cleaning existing data? While traditional imputation methods—including statistical interpolation and deep generative approaches—offer partial solutions, their efficacy varies significantly across tasks. This variation highlights persistent gaps in our understanding of the trade-offs between data quality, quantity, and imputation strategies in machine learning systems.

This study bridges these gaps through a systematic analysis of two distinct machine learning paradigms: supervised NLP tasks (NLP-SL) and deep reinforcement learning for traffic signal control (Signal-RL). It evaluates how varying corruption levels affect model performances, quantify the trade-offs of imputation strategies, and assess whether expanding datasets can counteract corruption effects. Experiments reveal universal patterns (such as diminishing returns from data quality improvements), paradigm-specific insights (such as DRL’s heightened sensitivity to noise), as well as task-specific observations (such as  the critical 30\% of data that determines performance in traffic signal control). This work provides actionable guidelines for designing robust systems in noisy environments, emphasizing smart data prioritization, imputation decisions, and noise-aware preprocessing. Key innovations include:
\begin{itemize}
    \item \textbf{Modeling Performance Degradation}: This study shows that performance degradation follows a diminishing-return pattern, well-captured by an exponential function, and reveal task-specific sensitivities.
    \item \textbf{Imputation Trade-offs}: This study demonstrates the boundary conditions where imputation is beneficial versus harmful, providing actionable guidelines for practitioners.
    \item \textbf{Data Quantity-Quality Trade-offs}: This study empirically shows that larger datasets can only partially offset corruption effects, with diminishing utility at high corruption levels.
\end{itemize}

By addressing these questions, our study advances the understanding of data corruption’s impact on machine learning and provides a framework for future research in robust model development.

\section{Related Work}

\subsection{Types of Data Corruption}

Data corruption in machine learning encompasses various forms, including missing data, noisy data, and adversarial perturbations. This study focuses on two major types of corruption: missing data and noisy data, which commonly occur in real-world scenarios. Missing data can result from sensor dropout, incomplete data collection, or non-response in surveys. Rubin’s classification categorizes missing data into three types: Missing Completely at Random (MCAR), Missing at Random (MAR), and Missing Not at Random (MNAR) \cite{Rubin1976}. In NLP, missing data often manifests as masked or unknown tokens, while in reinforcement learning, it appears as incomplete state observations. Data noise can affect both labels and features, with sources ranging from environmental factors to measurement errors. Noise is often categorized by its statistical distribution (e.g., Gaussian \cite{bishop2006pattern}/adversarial \cite{goodfellow2014explaining}) or types (e.g., additive/multiplicative \cite{moon2000mathematical}; label/feature \cite{song2022learning}). Noisy data significantly disrupts model learning, particularly when noise affects key features or labels critical to decision-making.

\subsection{Impacts of Data Corruption}

Prior research has empirically demonstrated the substantial impact of data corruption on model performance across learning paradigms. In the realm of language models, Joshi et al. (2020) \cite{joshi2020spanbert} found that missing rare tokens during pre-training led to incomplete token embeddings that limits the model’s ability to capture fine-grained semantics. Missing data reduces the available context, weakening learned representations and hindering downstream tasks such as summarization or question answering \cite{devlin2018bert, liu2019roberta}. Noisy data, particularly in large-scale corpora, introduces biases and degrades model robustness. Brown et al. (2020) \cite{brown2020language} highlighted that noisy training data increases the likelihood of generating biased or low-quality outputs, while filtering and robust training objectives mitigate such effects. Semantic noise, such as contradictory or irrelevant text, reduces the model’s ability to retain factual knowledge and generalize across tasks \cite{petroni2020kilt}. 

For reinforcement learning, data corruption affects state observations, which are critical for decision-making. Missing features reduce state informativeness, leading to suboptimal policies, particularly in partially observable environments (POMDPs) \cite{hausknecht2015deep}. Studies by Bai et al. (2019) \cite{bai2019model} demonstrated that missing features disrupt state transition dynamics, causing instability in model-free RL algorithms and inaccurate environment models in model-based RL . Pathak et al. (2017) showed that noisy features distort latent state representations, leading to poor decision-making in high-dimensional environments \cite{pathak2017curiosity}. Mnih et al. (2015) \cite{mnih2015human} found that Q-values fluctuate with noisy observations, resulting in unstable policies. Moreover, noise hampers transfer learning, as policies trained in corrupted environments fail to generalize to clean environments \cite{taylor2009transfer}. These studies collectively highlight how data corruption undermines performance in both paradigms, though through different mechanisms: contextual understanding and factual accuracy in language models versus compounding errors in sequential decision-making for reinforcement learning.

\subsection{Data Imputation Techniques}

Data imputation aims to recover missing information, and various strategies have been proposed. Emmanuel et al. \cite{emmanuel2021survey} provide a comprehensive review of missing data in machine learning, classifying imputation methods into several categories: simple imputation, regression-based imputation, hot-deck imputation, the Expectation–Maximization (EM) method, multiple imputation, and machine learning–inspired techniques such as k-nearest neighbors (KNN), support vector machines (SVM), decision trees, clustering-based imputation, and ensemble methods. In a more recent taxonomy, Zhou et al. \cite{zhou2024review} group imputation approaches into four main types: statistical, machine learning–based, neural network–based, and optimization-based methods.

\textbf{Statistical Methods}: These include simple approaches like mean, mode, or median imputation \cite{Schafer2002}, as well as more sophisticated techniques like regression imputation and multiple imputation \cite{Rubin1987}. While straightforward, these methods may introduce bias or underestimate variability.

\textbf{Machine Learning Methods}: Techniques such as K-Nearest Neighbors (KNN) imputation \cite{Troyanskaya2001}, decision tree-based imputation \cite{Breiman1984}, and Random Forest imputation \cite{Breiman2001} leverage patterns in data to predict missing values. These methods often outperform simple statistical approaches by capturing complex nonlinear relationships in the data.

\textbf{Deep Learning-based Imputation}: Autoencoders and Generative Adversarial Networks (GANs) have emerged as powerful tools for data imputation. Denoising autoencoders reconstruct inputs with noisy values \cite{Vincent2008}, while GANs generate plausible synthetic data to fill gaps \cite{Goodfellow2014}. Yuan et al. (2021) \cite{Yuan2021} used masked language models (e.g., BERT) to impute missing tokens in text data.

For many applications, particularly LLM pre-training, masking missing data is often sufficient. Che et al. (2018) showed that masking missing time-series data combined with recurrent neural networks (e.g., GRU) can effectively handle data missing \cite{che2018recurrent}. We summarize common imputation methods, their strengths and weaknesses, and use cases in Table \ref{tab:imputation_taxonomy}.

\begin{table}[ht]
\centering
\small  
\caption{Taxonomy of Imputation Techniques with Strengths, Weaknesses, and Use Cases}
\label{tab:imputation_taxonomy}
\setlength{\tabcolsep}{4pt}  
\begin{tabular}{>{\raggedright\arraybackslash}p{2.2cm}>{\raggedright\arraybackslash}p{2.5cm}>{\raggedright\arraybackslash}p{5cm}>{\raggedright\arraybackslash}p{3.5cm}}
\toprule
\textbf{Category} & \textbf{Method} & \textbf{Strengths/Weaknesses} & \textbf{Use Cases} \\
\midrule
\multirow{4}{*}{\parbox{2.2cm}{Statistical-based \cite{Schafer2002, Rubin1987}}} 
& Mean/Median/Mode & 
\begin{tabular}[t]{@{}l@{}} 
+ Simple, fast \\ 
– Ignores correlations, distorts variance 
\end{tabular} & 
Small datasets, MCAR data \\
\cmidrule{2-4}
& Maximum Likelihood & 
\begin{tabular}[t]{@{}l@{}} 
+ Handles MAR data well \\ 
– Computationally intensive 
\end{tabular} & 
Surveys, clinical trials \\
\cmidrule{2-4}
& Matrix Completion & 
\begin{tabular}[t]{@{}l@{}} 
+ Captures global structure \\ 
– Requires low-rank assumption 
\end{tabular} & 
Recommendation systems \\
\cmidrule{2-4}
& Bayesian Approach & 
\begin{tabular}[t]{@{}l@{}} 
+ Incorporates uncertainty \\ 
– Needs prior distributions 
\end{tabular} & 
Small datasets with domain knowledge \\
\midrule
\multirow{5}{*}{\parbox{2.2cm}{Machine Learning \cite{Troyanskaya2001, Breiman1984, Breiman2001}}} 
& Regression-based & 
\begin{tabular}[t]{@{}l@{}} 
+ Models feature relationships \\ 
– Assumes linearity 
\end{tabular} & 
Tabular data with correlations \\
\cmidrule{2-4}
& KNN-based & 
\begin{tabular}[t]{@{}l@{}} 
+ Non-parametric, local patterns \\ 
– Sensitive to $k$, scales poorly 
\end{tabular} & 
Small/medium datasets \\
\cmidrule{2-4}
& Tree-based & 
\begin{tabular}[t]{@{}l@{}} 
+ Handles nonlinearity \\ 
– Overfitting risk 
\end{tabular} & 
High-dimensional data \\
\cmidrule{2-4}
& SVM-based & 
\begin{tabular}[t]{@{}l@{}} 
+ Robust to outliers \\ 
– Kernel choice critical 
\end{tabular} & 
Nonlinear feature spaces \\
\cmidrule{2-4}
& Clustering-based & 
\begin{tabular}[t]{@{}l@{}} 
+ Group-aware imputation \\ 
– Depends on cluster quality 
\end{tabular} & 
Data with clear subgroups \\
\midrule
\multirow{5}{*}{\parbox{2.2cm}{Neural Network \cite{Vincent2008, Goodfellow2014, Yuan2021}}} 
& ANN-based & 
\begin{tabular}[t]{@{}l@{}} 
+ Flexible architectures \\ 
– Requires large data 
\end{tabular} & 
Complex feature interactions \\
\cmidrule{2-4}
& Flow-based & 
\begin{tabular}[t]{@{}l@{}} 
+ Exact density estimation \\ 
– High computational cost 
\end{tabular} & 
Generative tasks \\
\cmidrule{2-4}
& VAE-based & 
\begin{tabular}[t]{@{}l@{}} 
+ Handles uncertainty \\ 
– Blurry imputations 
\end{tabular} & 
Image/text incomplete data \\
\cmidrule{2-4}
& GAN-based & 
\begin{tabular}[t]{@{}l@{}} 
+ High-fidelity samples \\ 
– Training instability 
\end{tabular} & 
Media generation \\
\cmidrule{2-4}
& Diffusion-based & 
\begin{tabular}[t]{@{}l@{}} 
+ State-of-the-art quality \\ 
– Slow sampling 
\end{tabular} & 
High-stakes applications \\
\bottomrule
\end{tabular}
\end{table}
 
\textbf{Key Research Gaps}

While many studies address specific aspects of corrupted data handling, key questions remain unanswered:
\begin{itemize}
    \item \textbf{Impact of Data Corruption}: What is the quantitative relationship between data corruption ratio and model performance? Can this relationship be consistently modeled across tasks?

    \item \textbf{Effectiveness of imputation}: How do different imputation methods compare in mitigating the effects of missing data? Is it possible to fully restore the utility of corrupted data through imputation?

    \item \textbf{Trade-Off Between Data Quality and Quantity}: Can larger datasets compensate for data corruption? How much additional data is required to offset quality issues and does the marginal utility of additional data diminish with increasing corruption level?
\end{itemize}

This study aims to bridge the above gaps by evaluating the effects of data corruption on supervised and reinforcement learning tasks. This study explores the utility of data imputation methods and analyze the trade-off between data quality and quantity, providing insights to guide data collection and preprocessing strategies.

\section{Learning with corrupted data}

\subsection{Experiment Design}

Two experiments are designed---Natural Language Processing Supervised Learning (NLP-SL) and Traffic Signal Deep Reinforcement Learning (Signal-RL)---to investigate the impact of data corruption. These two vastly different experimental setups were chosen to demonstrate the generality and broad relevance of the research questions across a wide range of machine learning tasks. By selecting tasks with diverse characteristics, we aim to derive more general insights and uncover deeper connections between these seemingly unrelated domains.

\subsubsection*{NLP Supervised Learning (NLP-SL)}

The first experiment focuses on the GLUE benchmark tasks. Firstly, a BERT model is pre-trained using \href{https://huggingface.co/datasets/Salesforce/wikitext}{Wikitext} and \href{https://huggingface.co/datasets/bookcorpus/bookcorpus}{Bookcorpus} as the base model; then the base model is frozen. We add classification head on top of the base model to fine-tune it on eight tasks: \texttt{CoLA}, \texttt{SST-2}, \texttt{MRPC}, \texttt{STSB}, \texttt{QQP}, \texttt{MNLI}, \texttt{QNLI}, and \texttt{RTE} using \href{https://huggingface.co/datasets/nyu-mll/glue}{GLUE} dataset. The GLUE input sentences are corrupted by replacing certain words with the \texttt{[UNK]} token, while the labels $y$ remain uncorrupted. This type of data corruption is commonly encountered in natural language processing tasks. For instance, when digitizing text corpora, some words may become indistinguishable and are thus masked as unknown, whereas the associated labels are typically unaffected. Evaluating the amount of knowledge learned by a model remains a subjective challenge. In this experiment, performance is measured using classification-related accuracy metrics. Specifically:
\begin{itemize}
    \item The Matthews Correlation Coefficient (MCC) is used for CoLA;
    \item The average of Pearson and Spearman correlation coefficients is used for STSB;
    \item Test accuracy is used for the remaining tasks.
\end{itemize}
Baseline scores ($\frac{1}{2}$ for binary classification and $\frac{1}{3}$ for three-way classification) are subtracted from these metrics. The final model score is computed as the average of the scores across all tasks. To ensure consistency, hyperparameters such as the number of training epochs and learning rates are tuned using uncorrupted data and kept fixed for all subsequent experiments (see Table~\ref{tab:experiment_setup}). Model convergence for these experiments is illustrated in Figure~\ref{fig:convergence}. For clarity, this experiment is referred to as the ``Natural Language Processing---Supervised Learning (NLP-SL)'' experiment in the following sections. Note that the base model is used only for sentence embedding and is frozen; the goal is to test the model's ability to extract knowledge from corrupted data and store it in the feed-forward network.

For the NLP-SL task, two types of data corruption are introduced:
\begin{itemize}
\item \textbf{Data-missing}: Each word in the training samples has a probability $p$ of being replaced with a \texttt{[MASK]} token.
\item \textbf{Inserting-noise}: Each word in the training samples has a probability $p$ of being replaced with a randomly selected word from the vocabulary.
\end{itemize}

\subsubsection*{Traffic Signal Control Deep Reinforcement Learning (Signal-RL)}

\begin{figure}
    \centering
    \includegraphics[width=0.9\linewidth]{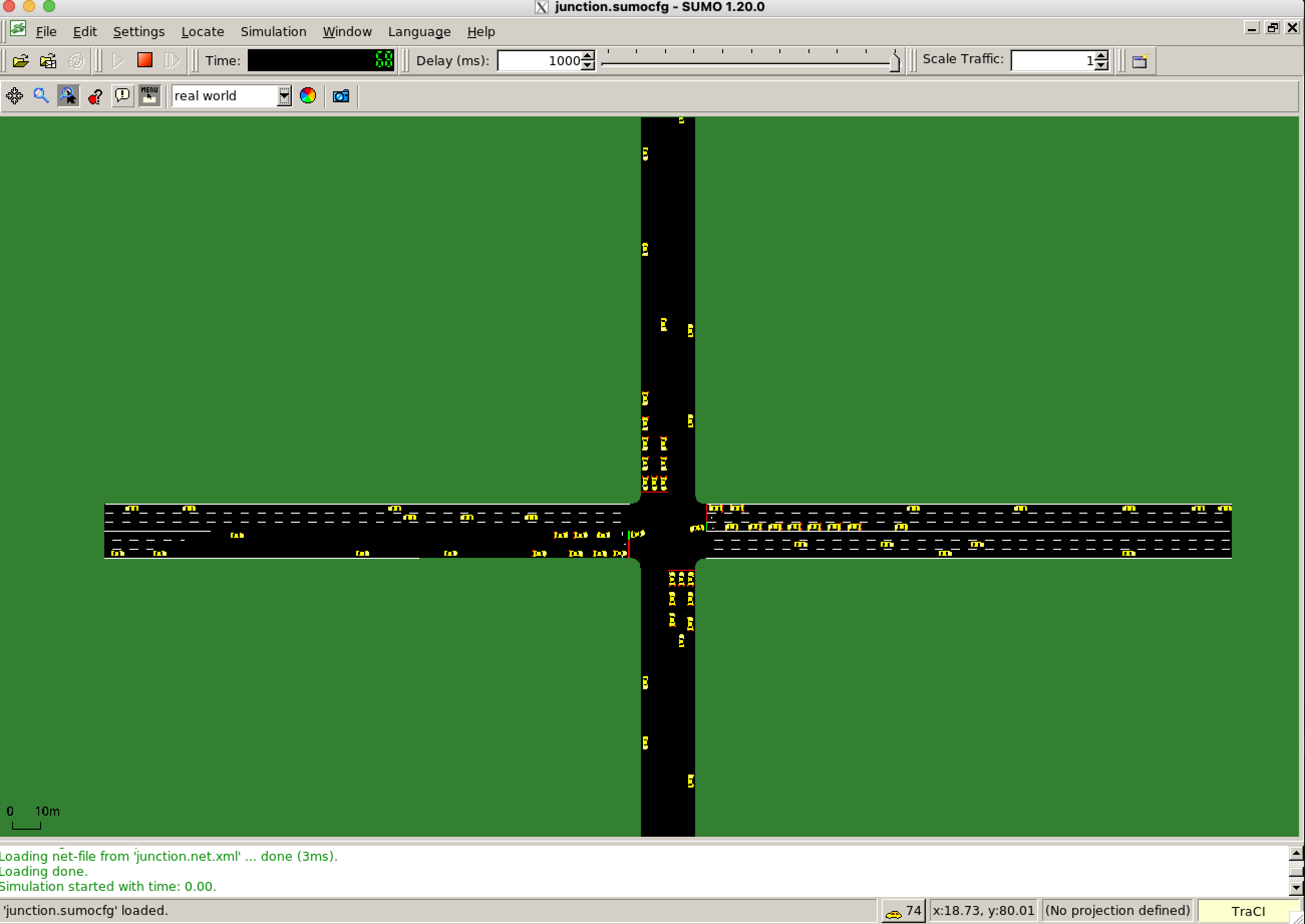}
    \caption{Simulation environment for Signal-RL experiment. The intersection comprises four approaches, each with three lanes. Traffic demand is generated based on a binomial distribution, with a left-turn:through:right-turn ratio of 1:3:2. The environment state is defined by road occupancy. Each simulation step corresponds to one second, and each episode spans one hour. The step-wise reward is defined in Equation~\ref{eq:signal_reward}.}
    \label{fig:sumo_illustration}
\end{figure}

The second experiment is a deep reinforcement learning (DRL) task. An isolated intersection environment is built using \href{https://eclipse.dev/sumo/}{SUMO}. The environment is illustrated in Figure \ref{fig:sumo_illustration}. The objective is to optimize the traffic signal at this intersection. The intersection consists of four approaches, each with three lanes. The traffic demand is generated using a binomial distribution, and the ratio of left-turn, through, and right-turn traffic demands is 1:3:2. The arrival rates are time-varying: East-West traffic demands follow a sine curve, while North-South demands follow a cosine curve within range $[0, \pi/2]$. The simulation step size is one second, and each episode has a horizon $H$ of one hour.

A Deep Q-Network (DQN) model (Table \ref{lst:dqn}) is used to learn the traffic signal control strategy. The state of the environment consists of road occupancy, the current signal phase, and the duration of the current signal phase, resulting in a state vector of dimension 965. A neural network with two hidden layers of sizes 256 and 48 is used to extract features. Layer normalization is applied, but no dropout layers are included. Standard techniques such as double networks and replay buffers are implemented. The action space is discrete, with four possible actions: East-West left turn, East-West through, North-South left turn, and North-South through. Each action in the simulation lasts for 6 seconds. The reward $r_t$ for each time step is defined as the number of queuing vehicles transformed according to Equation \ref{eq:signal_reward}. The queue length is divided by 80 to normalize step rewards approximately within the range of 0 to 1. The performance indicator for the model is the episode cumulative reward, $R$. Although the reward is accumulated over one hour, the process itself is infinite.

\begin{equation}
    \label{eq:signal_reward}
    R = \sum_{t=1}^{H}r_t = \sum_{t=1}^{H} -\frac{q_t - 80}{80} 
\end{equation}
where $q_t$ is the number of queuing vehicles at the intersection at time $t$; and the threshold for deciding whether a vehicle is stopped is $0.3 m/s$. $H$ is the horizon of one simulation episode.

\begin{figure}
    \centering
    \includegraphics[width=0.9\linewidth]{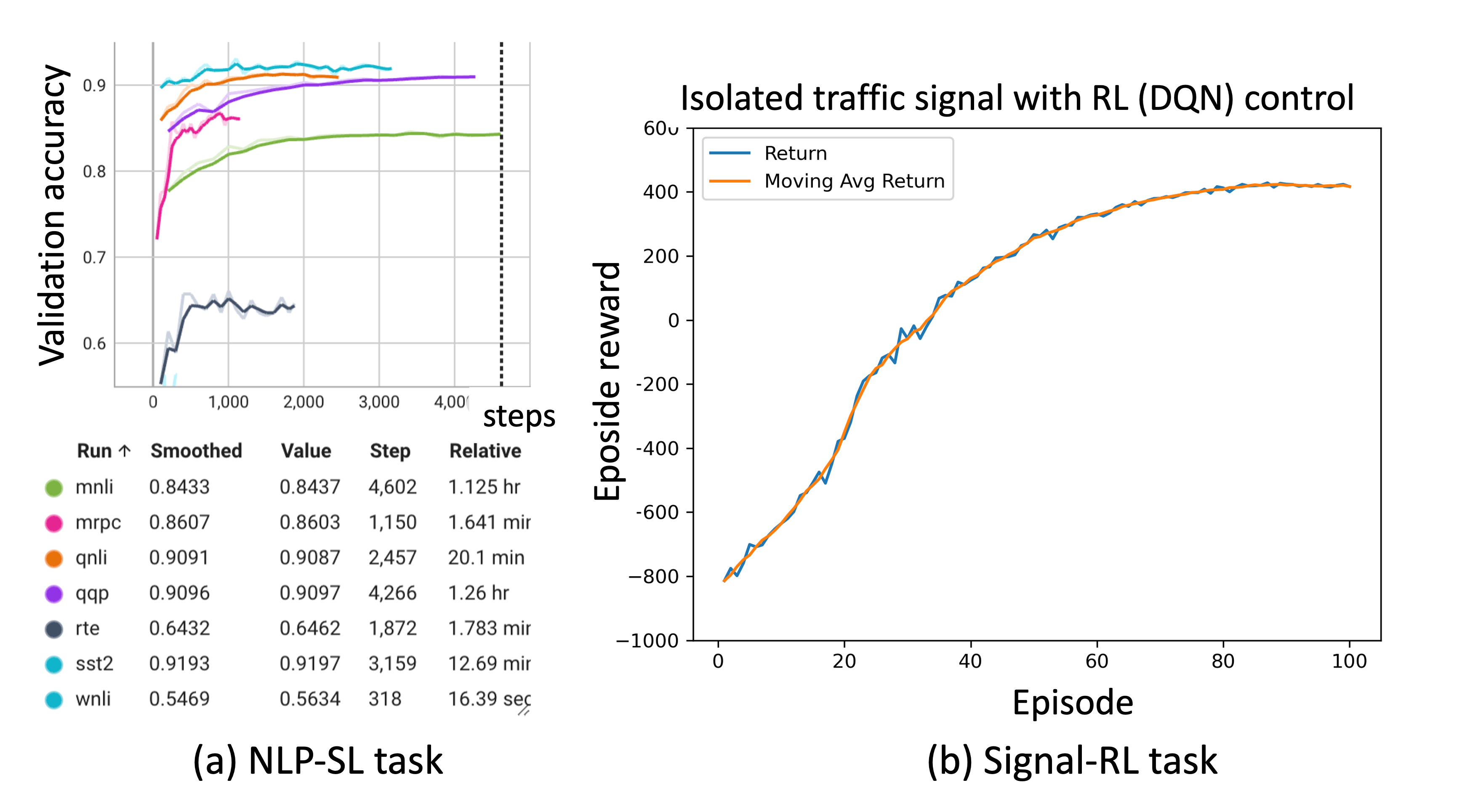}
    \caption{Model convergence in two experiments.
(a) NLP-SL experiment: The $x$-axis represents training steps, and the $y$-axis shows validation accuracy.
(b) Signal-RL experiment: The $x$-axis denotes training episodes, and the $y$-axis indicates episode reward.
Training configurations are detailed in Table~\ref{tab:experiment_setup}.}
    \label{fig:convergence}
\end{figure}

The model is trained using linearly decaying exploration ($\epsilon$) and learning rate ($lr$) for 80\% of the training time, after which the values were fixed. The initial and final $\epsilon$ values were 1.0 and 0.01, respectively, while the initial and final learning rates were 1e-3 and 1e-4, respectively. The DQN model convergence results are shown in Figure \ref{fig:convergence}(b). The overview of model, dataset, and training configurations for both NLP-SL and Signal-RL tasks are shown by Table \ref{tab:experiment_setup}.

For Signal-RL task, three types of data corruption are introduced: vehicle-missing, inserting-noise, and masking-region:
\begin{itemize}
    \item \textbf{Vehicle-missing}: Each vehicle is not detected with probability $p$. This scenario is relevant in Vehicle-to-Everything (V2X) environments, where roadside units detect vehicles’ presence through communication channels like DSRC \cite{tong2019artificial}. However, only a proportion of vehicles are equipped with onboard devices. This type of corruption is analogous to the data missing scenario in the NLP-SL experiment.
    \item \textbf{Inserting-noise}: Noise is added to the road occupancy state and rewards. Each road cell occupancy state has probability $p$ of being replaced with random binary value. This scenario is relevant in environments where road occupancies are detected using computer vision systems, which can introduce errors.
    \item \textbf{Masking-region}: This special type of corruption is specific to traffic signal settings. The simulation environment assumes a lane length of 400 meters. A masking-region ratio $p$ means that the farthest $400*p$ meters of each lane will be invisible to the model, simulating the range limitations of video cameras used for vehicle detection.
\end{itemize}

When investigating imputation methods, two common types of data-missing scenarios often encountered in practice are distinguished: 
\begin{itemize}
    \item \textbf{Exact imputation}: This scenario arises when the precise locations of missing data are known. A common example occurs in natural language processing (NLP), where missing words are explicitly marked with placeholders such as ``[UNK]''.
    \item \textbf{General imputation}: In this case, the locations of missing data are unknown. For instance, in the Signal-RL experiment, vehicles may go undetected, leaving it unclear which elements of the state vector are corrupted. Imputing data under such conditions requires checking all possible locations, potentially introducing significantly more noise compared to exact imputation.
\end{itemize}

\subsection{Observations}

\begin{figure}
    \centering
    \includegraphics[width=0.9\linewidth]{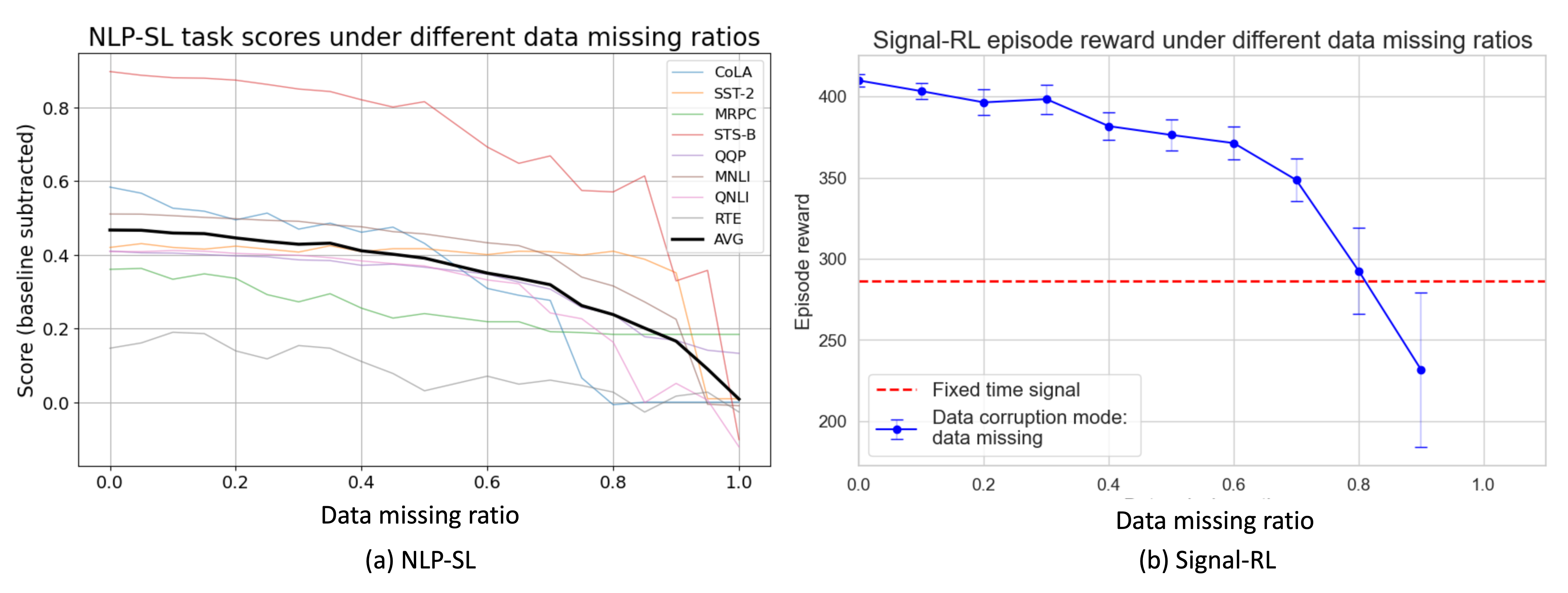}
    \caption{Model performance across varying data missing ratios. The $x$-axis indicates the data missing ratio, while the $y$-axis shows the model score—normalized accuracy for NLP-SL and normalized negative queue length for Signal-RL.
(a) NLP-SL experiment; (b) Signal-RL experiment. In both cases, model performance initially declines gradually, followed by a steeper drop and a sharp collapse as the missing ratio approaches 1.0. The RL-trained signal outperforms the fixed-timing signal when the data missing ratio is less than 0.8.}
    \label{fig:data_miss_relation}
\end{figure}

Figure~\ref{fig:data_miss_relation} illustrates the relationship between the data missing ratio and model performance. Both experiments (NLP-SL and Signal-RL) exhibit an initial gradual performance decline, followed by a steeper descent that culminates in a sharp performance drop as the data missing ratio approaches~1.0. For reference, the figure includes the performance of an optimized fixed-timing signal as a benchmark. The RL-trained signal outperforms the fixed-timing signal when the data missing ratio is below~0.8.

\begin{figure}
    \centering
    \includegraphics[width=0.9\linewidth]{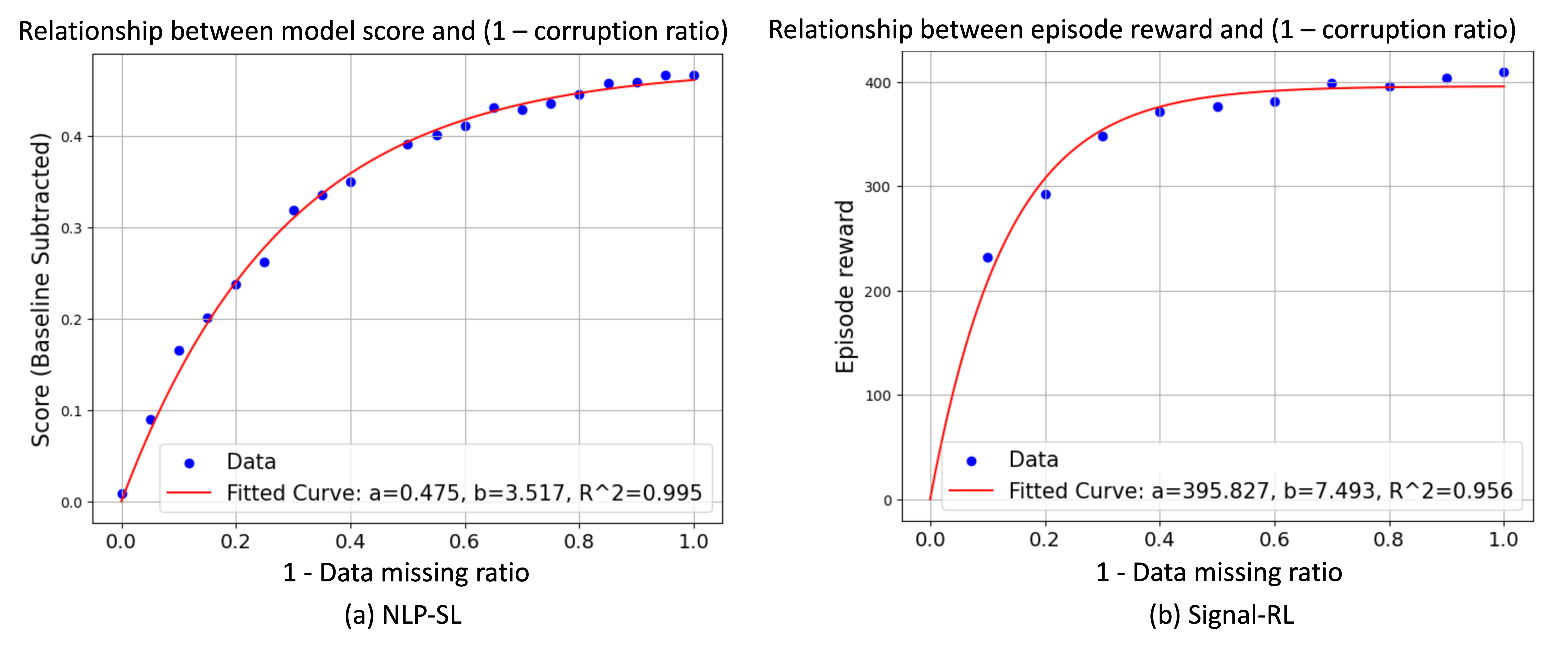}
    \caption{Relationship between model performance and data corruption ratio, fitted by Equation~\ref{eq:data_corruption_curve}. The $x$-axis represents $(1 - \text{corruption ratio})$, and the $y$-axis shows the model score—normalized accuracy for NLP-SL and normalized negative queue length for Signal-RL.
(a) NLP-SL experiment: Curve fitting parameters are $a = 0.475$, $\lambda = 3.517$, with $R^2 = 0.995$.
(b) Signal-RL experiment: Curve fitting parameters are $a = 395.8$, $\lambda = 7.493$, with $R^2 = 0.956$.
Both experiments exhibit diminishing returns as data quality improves. The fitted curves align well with the function defined in Equation~\ref{eq:data_corruption_curve}.}
    \label{fig:data_miss_curve_fit}
\end{figure}

Tables \ref{tab:data_missing_ratio_vs_score_nlp} and \ref{tab:missing_data_ratio_vs_score_signal} provide the detailed data for these experiments. In Figure \ref{fig:data_miss_curve_fit}, the $x$-axis is changed to represent (1 - \text{corruption ratio}) and fit the curve to the function in Equation \ref{eq:data_corruption_curve}. The fitted parameters for the NLP-SL experiment are: $a = 0.475$, $\lambda = 3.517$, where $\lambda$ represents the decay rate that controls the curve’s steepness. The goodness-of-fit analysis results in $R^2 = 0.995$, indicating that the exponential cumulative distribution function (CDF) is an excellent model for the observed performance. Similarly, for the Signal-RL experiment, the fitted curve parameters are: $a = 395.8$, $\lambda = 7.493$ with $R^2 = 0.956$.

This striking coincidence reveals an important and universal rule in machine learning: the diminishing return of data. The decay rate $b$ reflects the task’s nature, with the RL task demonstrating a much larger decay rate. This implies that RL tasks are more sensitive to data corruption compared to NLP tasks. An explanation for Equation \ref{eq:data_corruption_curve} is provided at the end of this section.

\begin{equation}
    \label{eq:data_corruption_curve}
    S = a (1 - e^{- \lambda (1 - p)})
\end{equation}
where parameter $a = \frac{S_0}{1 - e^{-\lambda}}$; and $S_0$ is the model score when corruption ratio $p$ = 0 (i.e., no corruption).

\begin{figure}
    \centering
    \includegraphics[width=0.9\linewidth]{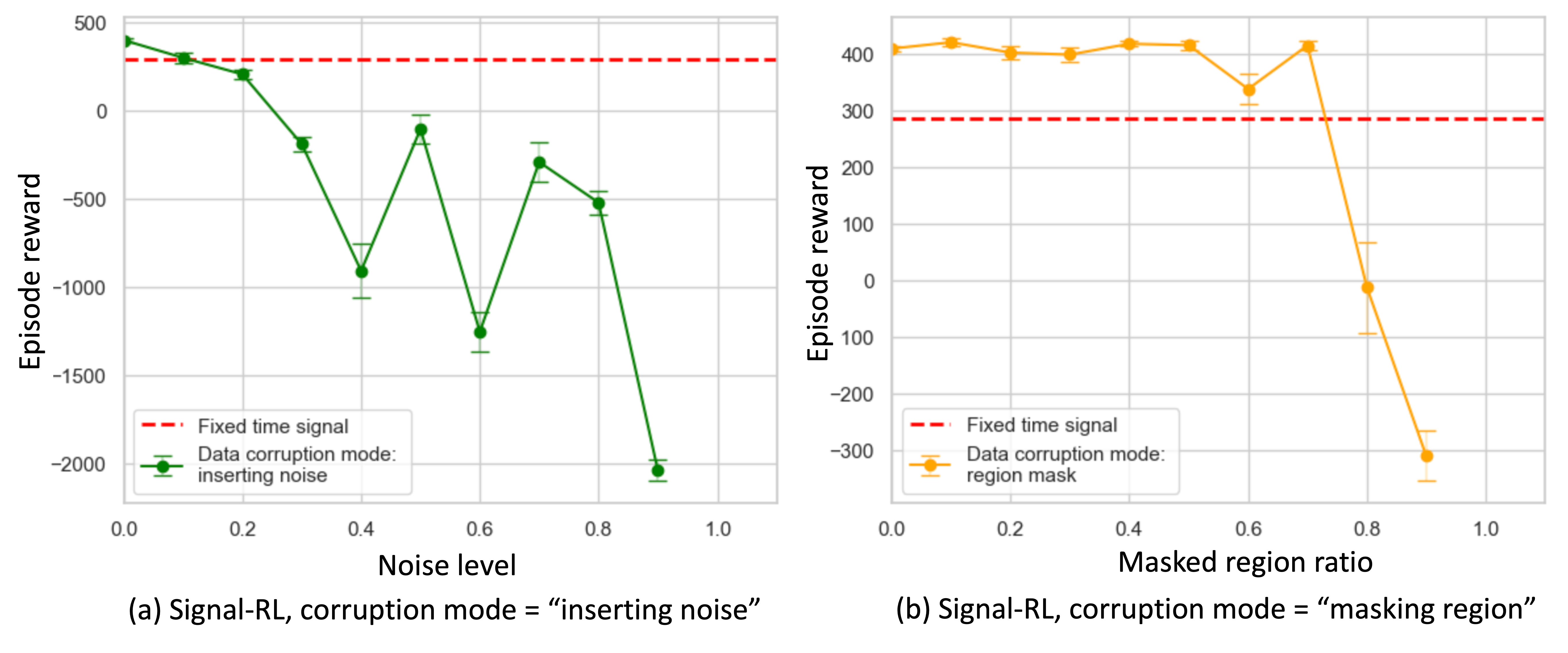}
    \caption{Signal-RL model performance under varying data corruption ratios.
(a) Noise insertion: The $x$-axis represents the noise level. Inserting noise is more detrimental than missing data, often leading to unstable model performance.
(b) Masked region: The $x$-axis denotes the masking-region ratio. A ratio of $p$ indicates that the farthest $400 \times p$ meters of each lane are hidden from the model, simulating the limited visual range of video-based vehicle detectors. Model performance remains stable even when up to 70\% of less critical data are removed.}
    \label{fig:data_corruption_effect_rl}
\end{figure}

\begin{figure}
    \centering
    \includegraphics[width=0.9\linewidth]{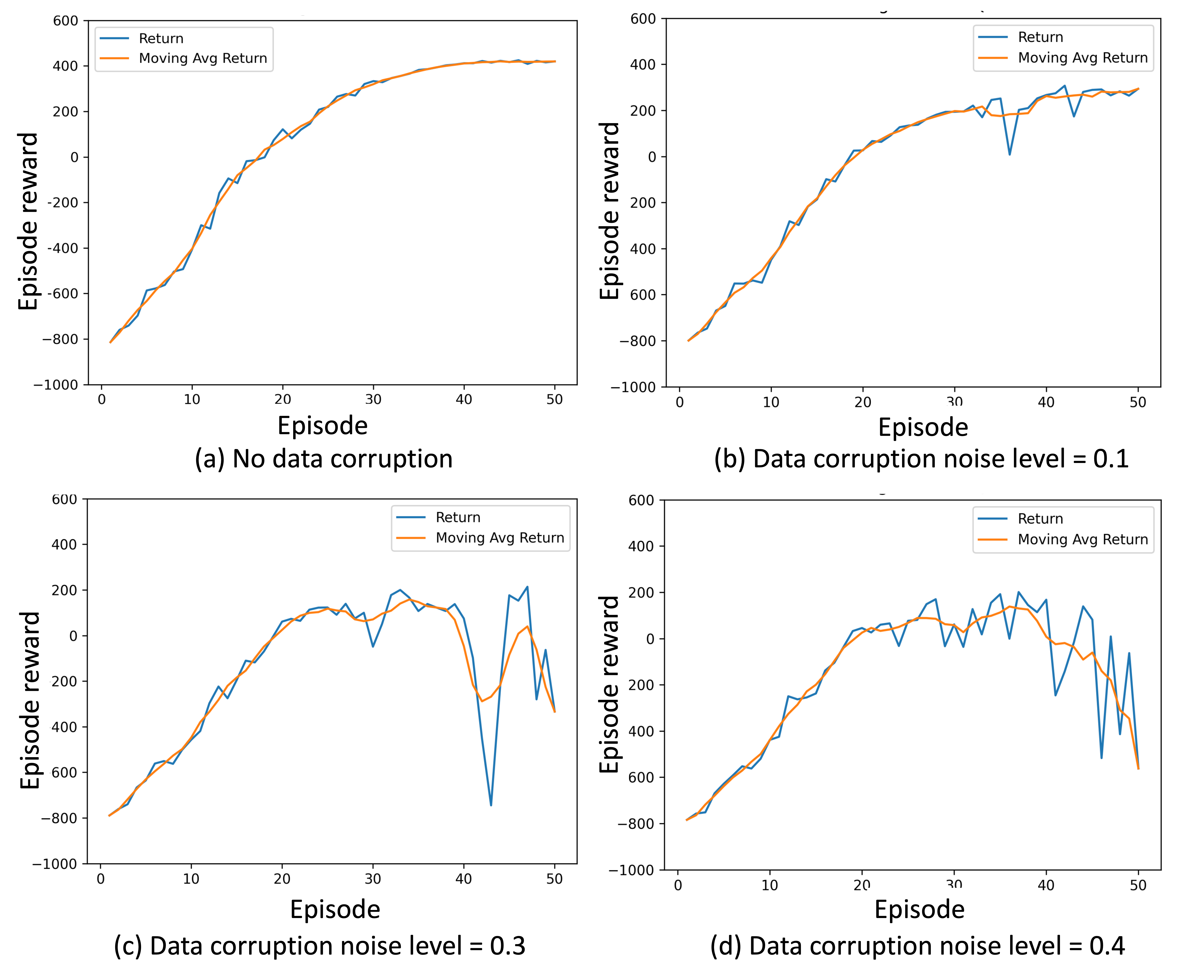}
    \caption{Training instability under noise-insertion corruption. The $x$-axis represents training episodes, and the $y$-axis shows episode return. From (a) to (d), the observation noise ratio increases from 0.0 to 0.4, leading to progressively more unstable training dynamics.}
    \label{fig:instabillity_when_insert_noise}
\end{figure}

The Signal-RL model scores under inserting-noise and masking-region types of data corruption are shown in Figure \ref{fig:data_corruption_effect_rl}. From Figure \ref{fig:data_corruption_effect_rl}(a), results show that inserting noise is significantly more detrimental than data-missing. The model’s performance deteriorates much more rapidly as the noise level increases, falling below the fixed-timing signal performance as soon as the noise level exceeds 10\%. Additionally, the model scores become unstable, as indicated by the oscillations in Figure \ref{fig:data_corruption_effect_rl}(a). The training process also becomes unstable, as shown in Figure \ref{fig:instabillity_when_insert_noise}.

Masking-region is a unique type of data-missing corruption. Information about vehicles farther away from the intersection is less critical. By gradually discarding less important data, it's observed that the model score only experiences a sharp decline when the masking ratio $p$ exceeds 0.7. This observation leads to an empirical rule: 30\% of the data is critical and determines the model’s performance, while the remaining 70\% can be missing without significantly affecting the model’s performance. The exact numbers may not hold for other tasks. And for many tasks (e.g. previous NLP tasks) this kind of screening is not feasible.

\begin{table}[h!]
\centering
\begin{tabular}{|c|c|c|c|c|c|}
\hline
\textbf{Data missing ratio} & 0.0   & 0.05  & 0.1   & 0.15  & 0.2   \\ \hline
\textbf{Model score}            & 0.4669 & 0.4663 & 0.4588 & 0.4572 & 0.4455 \\ \hline
\textbf{Data missing ratio} & 0.25  & 0.3   & 0.35  & 0.4   & 0.45  \\ \hline
\textbf{Model score}            & 0.4359 & 0.4283 & 0.4312 & 0.4106 & 0.4012 \\ \hline
\textbf{Data missing ratio} & 0.5   & 0.6   & 0.65  & 0.7   & 0.75  \\ \hline
\textbf{Model score}            & 0.3906 & 0.3501 & 0.3357 & 0.3186 & 0.2619 \\ \hline
\textbf{Data missing ratio} & 0.8   & 0.85  & 0.9   & 0.95  & 1.0   \\ \hline
\textbf{Model score}            & 0.2375 & 0.2008 & 0.1654 & 0.0897 & 0.0081 \\ \hline
\end{tabular}
\caption{NLP-SL model scores at different data missing ratios; plotted in Figure~\ref{fig:data_miss_relation}(a).}
\label{tab:data_missing_ratio_vs_score_nlp}
\end{table}

\begin{table}[h!]
\centering
\begin{tabular}{|c|c|c|c|c|c|}
\hline
\textbf{Data missing ratio} & 0.0 & 0.1 & 0.2 & 0.3 & 0.4 \\ \hline
\textbf{Model score mean}       & 409.86 & 403.30 & 396.40 & 398.40 & 381.74 \\ \hline
\textbf{Model score std}        & 3.83 & 5.04 & 7.92 & 8.97 & 8.44 \\ \hline
\end{tabular}
\vspace{0.5cm}
\begin{tabular}{|c|c|c|c|c|c|}
\hline
\textbf{Data missing ratio} & 0.5 & 0.6 & 0.7 & 0.8 & 0.9 \\ \hline
\textbf{Model score mean}       & 376.33 & 371.30 & 348.58 & 292.54 & 231.56 \\ \hline
\textbf{Model score std}        & 9.62 & 10.05 & 13.25 & 26.44 & 47.59 \\ \hline
\end{tabular}
\caption{Mean and standard deviation of Signal-RL model scores at different data missing ratios; plotted in Figure~\ref{fig:data_miss_relation}(b).}
\label{tab:missing_data_ratio_vs_score_signal}
\end{table}

\subsection{Explanation}

An explanation for the observed pattern in Figure \ref{fig:data_miss_curve_fit} and Equation \ref{eq:data_corruption_curve} is provided here. For both NLP and DRL tasks, the model relies on recognizing patterns (e.g., semantic patterns, queuing patterns) in the data. When the data is completely corrupted, the critical patterns necessary for performance are entirely lost. As corruption decreases, the model rapidly recovers key patterns, resulting in a steep improvement in performance. However, the marginal utility of additional clean data diminishes, leading to a saturation of performance.

First, some basic probability theory is introduced. The probabilities of observing rare events in a large number of trials converge to the Poisson distribution \cite{feller1991introduction}.  When $n$ (the number of trials) is large, $p^{\prime}$ (the probability of success per trial) is small and the expected number of successes $\lambda = n \cdot p^{\prime}$ is finite, the binomial distribution approximates the Poisson distribution:

\begin{equation}
    P(K = k) \approx \frac{\lambda^k e^{-\lambda}}{k!},
\end{equation}
where $\lambda = n \cdot p^{\prime}$ is the rate parameter, representing the expected number of successes. The exponential term $e^{-\lambda}$ in the Poisson distribution describes the probability of observing 0 successes.

 Next, the above theory is applied to our pattern recognition problem. In both experiments, the dataset size $n$ is large, making the discovery of a pattern from an individual sample a rare event. In our experiments, each pattern has an equal probability $p$ of being corrupted. Let $x = 1 - p$. A pattern can be recovered multiple times from samples in the dataset with probability $P(K = k)$. The probability of failing to recover such a pattern with corruption level $p$ is $e^{- \lambda x}$, where $\lambda$ is the pattern appearance rate given no corruption. As $x$ increases, the probability of failing to recover a pattern have derivative $- \lambda e^{ - \lambda x}$. In other words, the probability of identifying such a pattern increase by $\lambda e^{- \lambda x}$ marginally. Suppose that model socre $S$ is proportional to the number of patterns identified. Then, as $x$ increases, the rate of $S$ is proportional to $\lambda e^{- \lambda x}$. This leads to Eqn~\ref{eq:system_evolving} where $a$ is coefficient for this linear relation. Its solution corresponds to Eqn \ref{eq:data_corruption_curve}. It can be shown that $\lambda e^{- \lambda x} = \lambda (a - S)$, so Eqn \ref{eq:system_evolving} actually describes a dynamic system where the rate of change in performance depends on the difference between the system’s current performance and its limit.

\begin{equation}
    \label{eq:system_evolving}
    \frac{dS}{dx} = a \lambda e^{- \lambda x}
\end{equation}
where $S$ is the model score; $x = 1 - p$ and $p$ is corruption rate; $b$ is pattern appearance rate; $a$ is the model score when there is no corruption.

\section{Effectiveness of Data Imputation}

This section is aimed at assessing whether and how different imputation strategies mitigate the impact of missing data. It is noted that the decision to impute missing data involves a trade-off between recovering missing information and potentially introducing noise.

\subsection{Experiment Design}

These imputation methods will be evaluated under varying data-missing ratios. There are various types of imputation methods, as reviewed in the literature. However, evaluating the effectiveness of these algorithms is non-trivial, and their accuracies are not adjustable parameters. For this study, where the focus is on the trade-off between data-missing and noise, it is crucial to control the accuracy of imputation. To achieve this, an artificial imputation method - “inserting-noise” - is proposed. For this method, accurate words (for NLP-SL) or state elements (for Signal-RL) are randomly (with probability $q$) replaced with random values, allowing us to control the imputation method noise level $q$. The example code for NLP-SL is provided below. The function \texttt{perturb\_sentence} is responsible for adding missing-type corruption to clean data and carrying out the artificial imputation. Later in this section, traditional imputation methods are also evaluated.

Let $S(p)$ represent the model score when the data-missing ratio is $p$ and no imputation is applied. Let $\tilde{S}(p, q)$ represent the model score when the data-missing ratio is $p$ and an imputation method with noise level $q$ is used to preprocess the data before training. The \textit{imputation advantage}, $A(p, q)$, is defined as:

\begin{equation}
    \label{eq:imputation advantage}
    A(p, q) = \tilde{S}(p, q) - S(p)
\end{equation}

\begin{figure}
    \centering
    \includegraphics[width=0.9\linewidth]{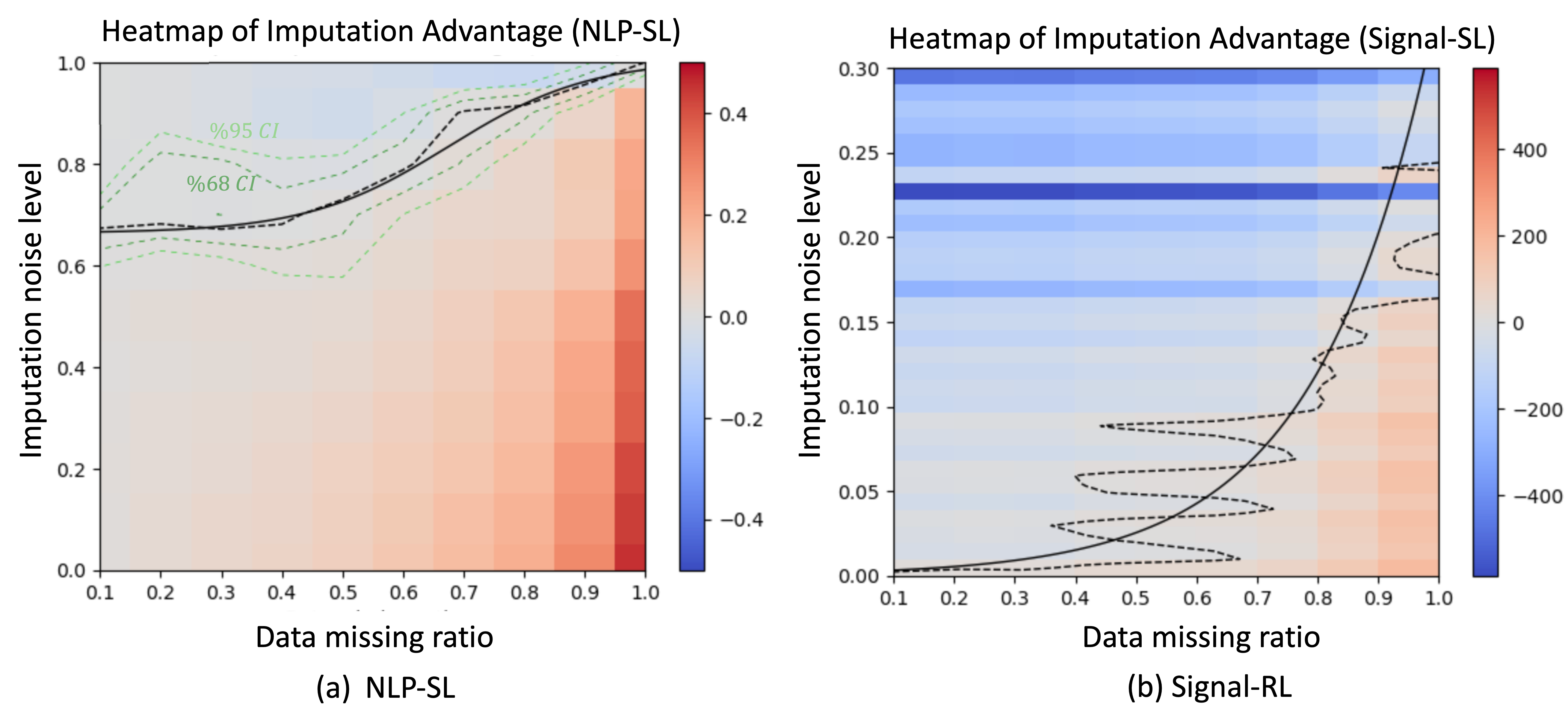}
    \caption{Heatmap of imputation advantage. The $x$-axis denotes the data missing ratio $p$, and the $y$-axis represents the imputation noise level $q$. The black dashed line indicates the decision boundary separating regions where imputation is beneficial from those where it is detrimental. Black solid line is the fitted curve of decision boundary. The green and lime green dashed lines denote the 68\% and 95\% confidence intervals.
(a) NLP-SL experiment: NLP-SL is a noise-insensitive task. The decision boundary lies above the diagonal and is well approximated by a logistic function.
(b) Signal-RL experiment: Signal-RL exemplifies a noise-sensitive task. The contour line appears below the diagonal, more irregular in shape, and fits an exponential curve starting from (0, 0) and intersecting the line $p = 1$.}
    \label{fig:imputation_advantage_heatmaps}
\end{figure}

$A(p, q)$ quantifies the improvement (or harm if negative) caused by imputation relative to no-imputation. Figure \ref{fig:imputation_advantage_heatmaps} shows the heatmap of imputation advantage for both experiments. 

\subsection{Observations}

Several interesting observations emerge from the heatmaps. First, the Signal-RL task demonstrates significantly greater sensitivity to imputation noise. This can be explained by the fact that sequential decision-making is inherently more noise-sensitive, combined with this task's use of ``general imputation''. In the heatmaps, red regions indicate where imputation improves model performance, while blue regions show where imputation is detrimental. These visualizations clearly reveal the trade-off between data missingness and noise introduction, with two distinct regions being identifiable:
\begin{itemize}
    \item \textbf{Imputation advantageous corner}: This region is located in the lower right corner of the heatmap, where the data-missing ratio is high, and the imputation noise level is low. Accurate imputation in this region restores critical information, leading to significant improvements in model performance.
    \item \textbf{Imputation disadvantageous edge}: This region is near the edge where $q$ = 1. When the imputation noise level approaches 1.0, the noise introduced during imputation overwhelms the model, leading to performance degradation. Interestingly, the greatest harm occurs when the data-missing ratio is around $p$ = 0.6.
\end{itemize}

Additionally, the black dashed contour line corresponding to $A(p, q) = 0$ is overlaid on the heatmap, indicating the decision boundary between regions where imputation is advantageous versus disadvantageous. The black solid line shows the fitted decision boundary. The green and lime green dashed lines denote the 68\% and 95\% confidence intervals, corresponding to $\pm 1$ and $\pm 1.96$ standard errors, respectively. A few notable differences between the Signal-RL and NLP-SL tasks can be observed:
\begin{itemize}
    \item \textbf{Signal-RL Decision Boundary}: The contour curve for the Signal-RL task lies much lower and is shifted to the right compared to the NLP-SL task. Moreover, the contour curve for Signal-RL is more ragged and fits to an exponential curve that starts at (0, 0) and intersects the line $p = 1$.
    \item \textbf{NLP-SL Decision Boundary}: The contour line for the NLP-SL task is smoother and fits well to a logistic function. When $p$ in range $[0, 0.4]$, the decision boundary is relatively stable and remains around $q = 0.68$. For $p$ in range $[0.4, 1.0]$, the contour line transitions into a sigmoid curve, with its midpoint around $p = 0.7$. This gradual transition reflects the trade-off between recovering critical information through accurate imputation and introducing ambiguities (e.g., incorrect word predictions) through noisy imputation.
\end{itemize}

\begin{figure}
    \centering
    \includegraphics[width=0.5\linewidth]{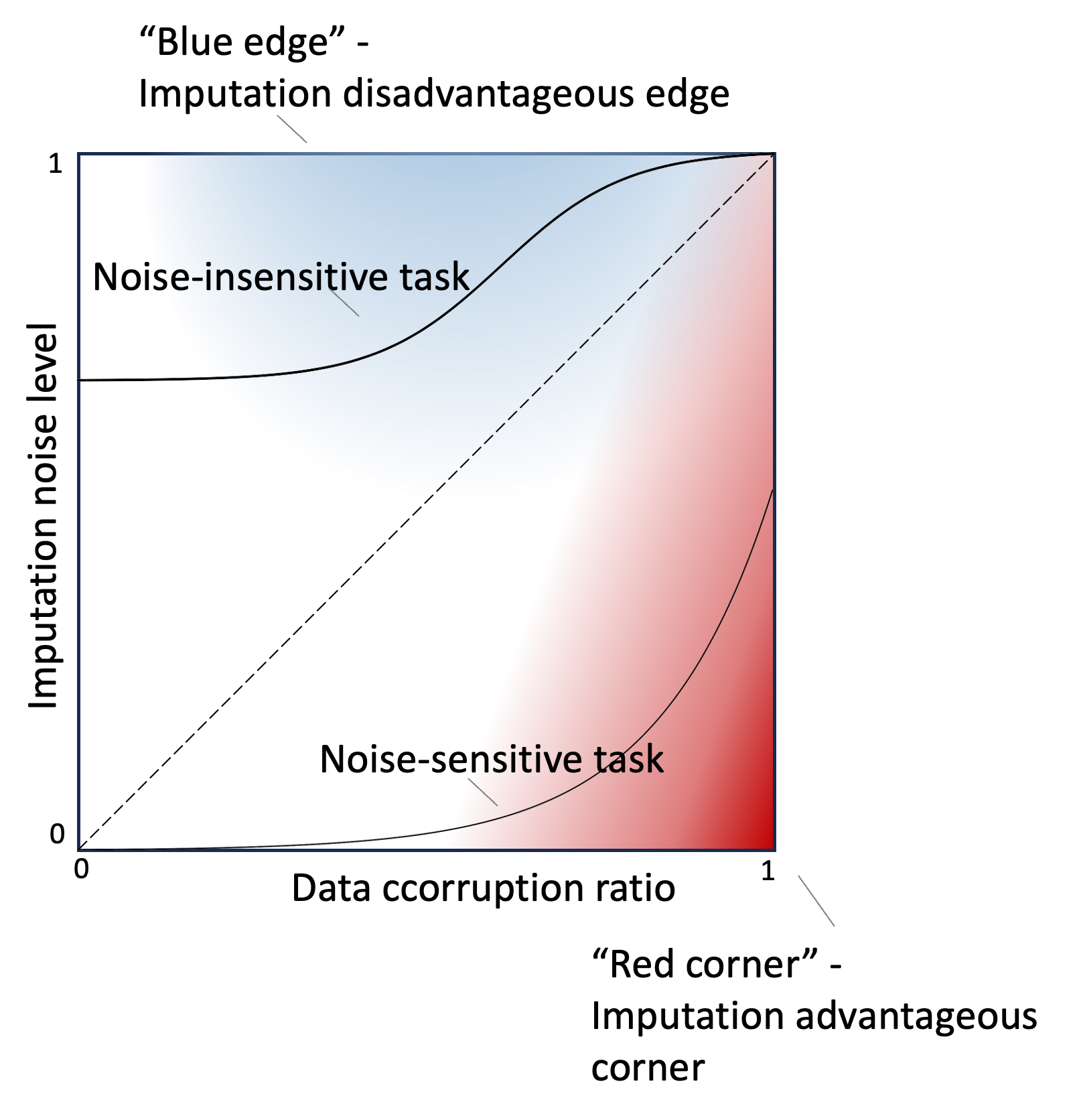}
    \caption{Illustration of imputation advantage pattern. The $x$-axis denotes the data missing ratio $p$, and the $y$-axis represents the imputation noise level $q$. ``Imputation advantageous corner'' and ``Imputation disadvantageous edge'' are areas where imputation advantage and disadvantage concentrate respectively. Task with decision boundary below diagonal line is called ``noise-sensitive'' and its decision boundary is fitted by exponential function. Task with decision boundary above diagonal line is called ``noise-insensitive'' and its decision boundary is fitted by Logistic function.}
    \label{fig:imputation_advantage_pattern}
\end{figure}

The sigmoid shape of the NLP-SL contour line reflects a smoother transition between advantageous and disadvantageous regions as imputation noise increases, whereas the Signal-RL task exhibits an exponential drop-off in performance due to the compounding effect of errors in sequential decision-making. tasks can be classified based on their sensitivity to noise:
\begin{itemize}
    \item \textbf{Noise-sensitive Tasks}: Tasks with contour curves below the diagonal (e.g., Signal-RL) are highly sensitive to noise, showing sharp performance degradation as imputation noise increases.
    \item \textbf{Noise-insensitvie Tasks}: Tasks with contour curves above the diagonal (e.g., NLP-SL) are more robust to imputation noise.
\end{itemize}
These observations are summarized in the Figure \ref{fig:imputation_advantage_pattern}. 

\begin{figure}
    \centering
    \includegraphics[width=0.9\linewidth]{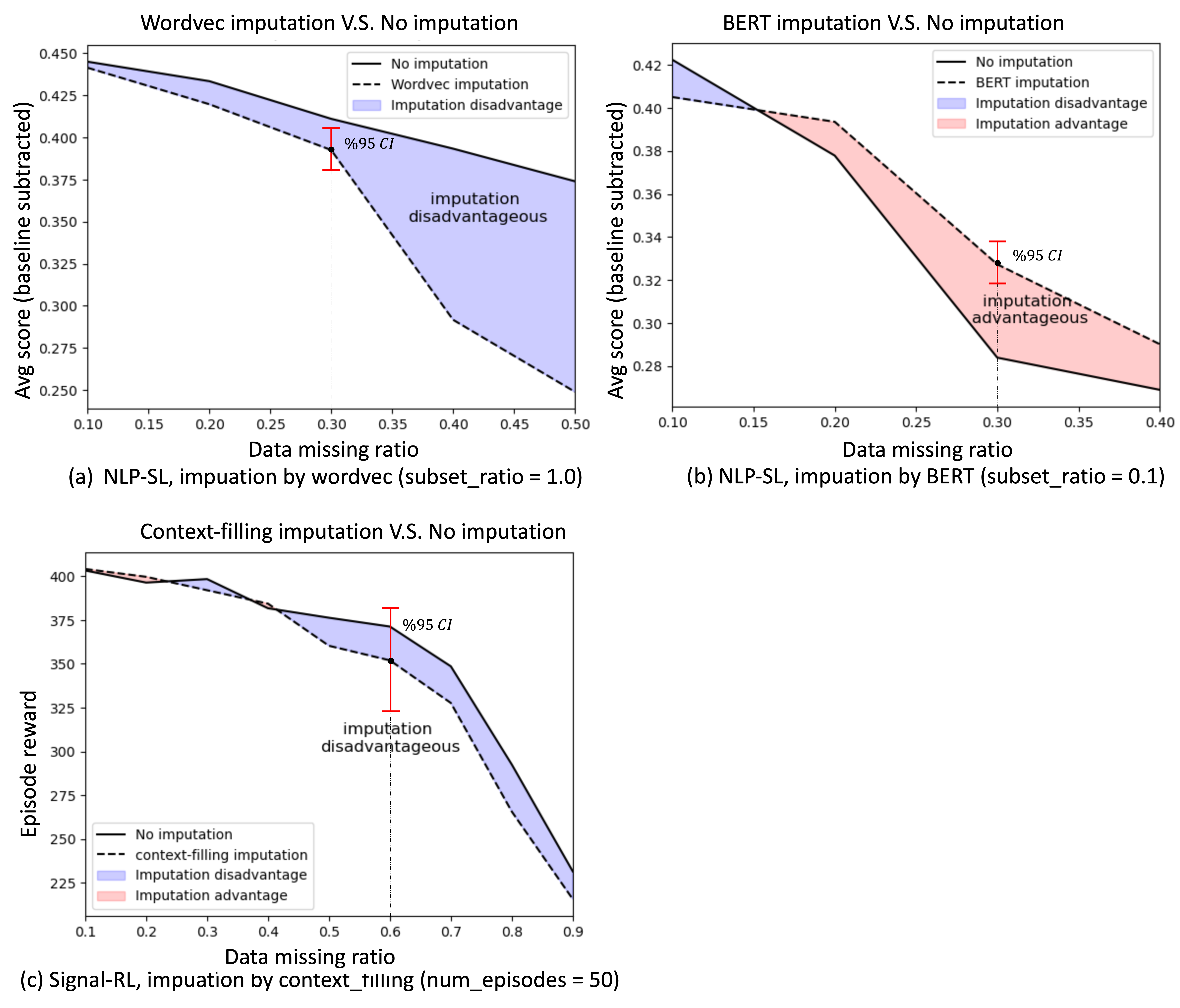}
    \caption{Effectiveness of alternative imputation methods. The $x$-axis represents the data missing ratio, and the $y$-axis indicates the model score (normalized accuracy). Solid lines denote the no-imputation baseline, while dashed lines represent model performance with imputation.
(a) Imputation using word vectors (NLP-SL task): This method introduces substantial noise, resulting in a significant performance drop compared to the no-imputation baseline.  
(b) Imputation using BERT (NLP-SL task): BERT-based imputation effectively recovers informative content useful for classification while introducing considerably less noise, leading to improved performance over the no-imputation baseline. At  \texttt{$p$ = 0.3}, \texttt{model score difference = +0.046},  \texttt{std error} = 0.0043,  \texttt{$z$-score = 10.65}; the difference is significant.
(c) Context-filling imputation (Signal-RL task): This method shows no clear advantage over the no-imputation.  At  \texttt{$p$ = 0.6},  \texttt{model score difference = -22},  \texttt{std error = 15},  \texttt{$z$-score = -1.48},  \texttt{$p$-value = 0.14}; the difference is not significant.}
    \label{fig:imputation_advantage_other_methods}
\end{figure}

Additional experiments were conducted using other common imputation methods. For NLP-SL, two imputation methods are introduced: “wordvec” and “BERT”. The wordvec imputation uses context word vector embeddings (GloVe) and cosine similarity to impute missing words. The BERT method leverages the pre-trained \texttt{bert-base-uncased} model for imputation. These two imputation methods are examples of ``exact imputation''. The results of these experiments are shown in Figure \ref{fig:imputation_advantage_other_methods} (a) and (b). To speed up preprocessing, a subset ratio of 0.1 was used for BERT imputation. 

The model score for the NLP-SL task shows an increasing standard error, rising from 0.0015 to 0.0227 as the data corruption ratio increases from 0.1 to 1.0. Imputation using Word2Vec leads to a significant performance decline relative to the no-imputation baseline (Figure \ref{fig:imputation_advantage_other_methods}(a)), indicating that this approach introduces more noise than it mitigates. BERT-based imputation effectively reconstructs informative content for classification while introducing minimal additional noise, resulting in better performance compared to the no-imputation condition (Figure \ref{fig:imputation_advantage_other_methods}(b)). At a noise-missing ratio of 0.3, the model score difference between BERT-based imputation and the no-imputation baseline is +0.046, with a standard error of 0.0043, yielding a $z$-score of 10.65. This difference is statistically significant. An important observation is that common imputation methods do not maintain a fixed accuracy level as the missing data ratio $p$ varies. Instead, their performance corresponds to curves, rather than horizontal lines, on the advantage heatmap.

\begin{figure}
    \centering
    \includegraphics[width=0.9\linewidth]{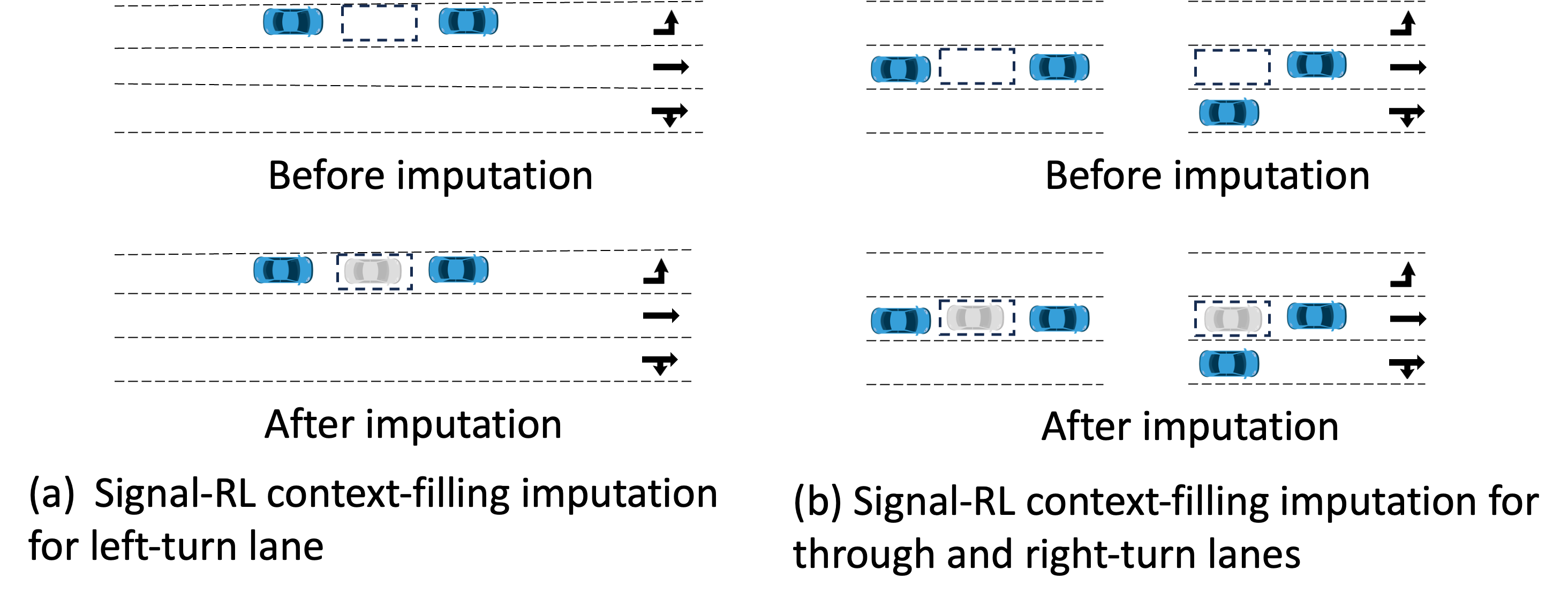}
    \caption{Illustration of context-filling imputation in the Signal-RL experiment. This represents a type of ``general imputation'', where each element of the state vector is evaluated for potential correction. Road occupancy cells with a value of zero are imputed as one if sufficient surrounding vehicles are detected.}
    \label{fig:imputation_signal_illustration}
\end{figure}

For Signal-RL tasks, an imputation method called ``context-filling'' is proposed. In this approach, road occupancy cells with a value of zero are imputed as one if sufficient surrounding vehicles are detected, as illustrated in Figure \ref{fig:imputation_signal_illustration}. This imputation method is an example of ``general imputation'', where each element of the state vector is evaluated for potential correction. Model performance with context-filling imputation, compared to the no-imputation baseline, is presented in Figure~\ref{fig:imputation_advantage_other_methods}(c). The model score exhibits substantial variability, with standard errors ranging from 4.8 to 47.6 as the corruption level increases from 0.1 to 0.9. Context-filling shows no clear advantage and in fact performs slightly worse in our results. At a data-missing ratio of 0.6, the model score between context-filling and no imputation is 22, with a standard error of 15. The $z$-score is 1.48 and the $p$-value is 0.14. This difference is not substantial given this high performance variance observed in the Signal-RL experiments.

\section{Effectiveness of Enlarging Dataset}

\subsection{Experiment Design}

The aim of this section is to evaluate the effectiveness of enlarging the dataset and to quantify how much additional data is needed to offset the effects of data corruption. For the NLP-SL experiment, the model is tested on datasets of different sizes and varying data corruption ratios. The variable “subset ratio” represents the proportion of the GLUE dataset used for fine-tuning. The results are shown in Figure \ref{fig:enlarging_dataset} (a). The Signal-RL task experiments with different numbers of training episodes and data-missing ratios, as shown in Figure \ref{fig:enlarging_dataset} (b).

\subsection{Observations}

\begin{figure}
    \centering
    \includegraphics[width=0.9\linewidth]{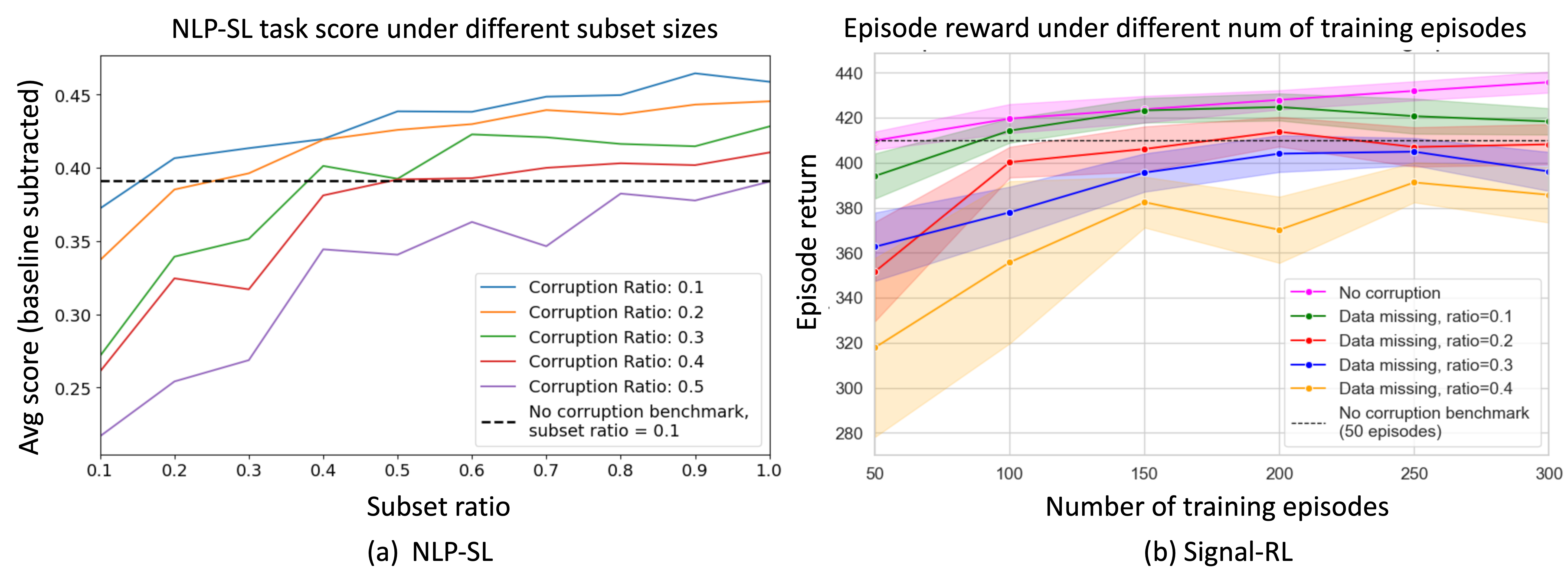}
    \caption{Effectiveness of enlarging training dataset. (a) NLP-SL. The $x$-axis represents subuset ratio, and the $y$-axis indicates model score. (b) Signal-RL. The $x$-axis represents number of training episodes, and the $y$-axis indicates model score. Data corruption leads to a decline in model performance, which cannot be fully recovered by increasing the sample size. To achieve the same level of performance (if achievable by the corrupted model), the number of samples, and hence the training time, required increases exponentially with the data corruption level.}
    \label{fig:enlarging_dataset}
\end{figure}

As shown in the figure, as the dataset size increases, model performance converges. However, the results show that data corruption leads to a decline in model performance that cannot be fully recovered by increasing the sample size. When corruption ratio $p$ is in the range [0, 0.4], the performance decline is nearly linear with respect to $p$ (Figure \ref{fig:enlarging_dataset} (a)). For larger $p$, the model’s performance drops sharply to near zero (Figure \ref{fig:data_miss_relation}).

This behavior is characteristic of an exponential function: for $e^x = 1 + x + \dots$, the linear term dominates when $x$ is small. Hence, it's concluded that the performance drop increases approximately exponentially as the data corruption ratio increases. In addition, data corruption also hampers learning efficiency. To achieve the same level of performance (if achievable at all for a corrupted model), the number of samples—and therefore the training time—required increases exponentially with the data corruption level. This is illustrated by the dashed benchmark line in Figure \ref{fig:enlarging_dataset} (b). Quantitative curves showing the relationship between data quality and the required amount of data provide practical insights into data collection strategies.

\section{Conclusions}
This study explored the impact of data corruption, including missing and noisy data, on deep learning performance across two distinct domains: supervised learning with NLP tasks (NLP-SL) and deep reinforcement learning for traffic signal optimization (Signal-RL). Our experiments aimed to provide insights into the relationship between data quality and model performance, the trade-offs of imputation strategies, and the effectiveness of increasing data quantity as a remedy for data corruption.

\textbf{Key Findings}
\begin{enumerate}
    \item \textbf{Diminishing Returns in Data Quality Improvement}:
Both NLP-SL and Signal-RL experiments revealed that model performance follows a diminishing return curve as data corruption decreases. The relationship between the model score $S$ and data corruption level $p$ is well-modeled by the function:
\[
    S = a (1 - e^{- \lambda (1 - p)})
\]
where parameter $a = \frac{S_0}{1 - e^{-\lambda}}$; and $S_0$ is the model score when corruption ratio $p$ = 0.
This universal trend emphasizes the importance of balancing data quality and preprocessing efforts.

    \item \textbf{Data Noise is More Detrimental than Missing Data}:
    Our results demonstrate that noisy data is significantly more detrimental than missing data, leading to faster performance degradation and increased training instability. This was particularly evident in reinforcement learning task, where inserting noise caused substantial fluctuations in both training and policy stability.

    \item \textbf{Trade-offs in Data Imputation}:
    Imputation methods can restore critical information for missing data but introduce a trade-off by potentially adding noise, resulting in a trade-off. The decision to impute depends on the imputation accuracy, the corruption ratio, and the nature of the task. The imputation advantage heatmap highlights two key regions:
    \begin{itemize}
        \item Imputation Advantageous Corner: A region where accurate imputation significantly boosts model performance.
        \item Imputation Disadvantageous Edge: A region where imputation noise outweighs its benefits, harming model performance.
    \end{itemize}

    \item \textbf{Two Types of Tasks Identified}:
    Tasks are classified into two categories based on their sensitivity to noise:
    \begin{itemize}
        \item Noise-insensitive tasks: These tasks exhibit gradual performance degradation, with decision boundaries on the heatmap that can be effectively modeled using a sigmoid curve.
        \item Noise-sensitive tasks: These tasks exhibit sharp performance drops, with decision boundaries closely approximated by an exponential curve. This behavior is typical in deep reinforcement learning tasks. When only ``general imputation'' is available—as opposed to ``exact imputation''—the sensitivity to noise tends to be further amplified.
    \end{itemize}

    \item \textbf{Limits of Enlarging Datasets}:
    Increasing the dataset size partially mitigates the effects of data corruption but cannot fully recover the lost performance, especially under high noisy levels. Enlarging datasets does not entirely offset the detrimental effects of noisy data. The analysis showed that the number of samples required to achieve a certain performance level increases exponentially with the corruption ratio, confirming the exponential nature of the trade-off between data quality and quantity.

    \item \textbf{Impact of Data Corruption on Learning Efficiency}:
    Data missing hampers learning efficiency. To achieve the same performance level (if at all possible), the number of required samples—and hence training time—increases exponentially with the data missing level.

    \item \textbf{Empirical Rule on Data Importance}:
     For traffic signal control tasks, approximately 30\% of the data is critical for determining model performance, while the remaining 70\% can be lost with minimal impact on performance. This observation provides practical guidance for prioritizing efforts in data collection and preprocessing. Note that the exact number may not apply to other tasks. And for many task this kind of screening is not feasible.
\end{enumerate}

\textbf{Implications}

The insights from this study provide direct implications for designing robust machine learning systems, benefiting both practitioners and researchers. Imputation strategies should be tailored to task-specific sensitivity, leveraging the identified boundary conditions between beneficial and harmful imputation.  For noise-sensitive tasks, such as reinforcement learning, conservative imputation approaches and stricter data quality controls are essential to maintain model stability and performance. Practitioners should prioritize accurate data collection and preprocessing for the critical subset identified as having the most significant impact on performance. Specifically, in scenarios with limited resources, efforts should focus on minimizing noise in key data elements rather than indiscriminately expanding the dataset. 

\textbf{Future Work}

Although this study evaluates two distinct tasks, the generalizability of its findings to other domains remains uncertain. The current study employs straightforward corruption methods, whereas real-world data corruption can be more complex, involving correlated or structured patterns of corruption not accounted for in the current experiments. Additionally, the artificial method of controlling imputation accuracy by inserting random noise provides useful insights into theoretical trade-offs but may not fully capture the nuances of practical imputation techniques. Moreover, although standard imputation methods were investigated, more advanced and sophisticated approaches remain underexplored, leaving their potential effectiveness unclear.

This study opens up several avenues for future research. First, the observed patterns and empirical rules should be validated across broader datasets and additional machine learning tasks, such as computer vision and time-series forecasting. While the core principles—exponential performance decay, imputation decision boundaries—are hypothesized to generalize, their manifestations will likely depend on domain-specific factors. For instance, spatial correlations in CV or temporal dependencies in time-series data may influence how corruption impacts model performance. Notably, CV tasks may exhibit greater noise resilience compared to NLP, as pixel-level redundancies in images can mitigate localized corruption, whereas natural language is heavily compressed. Second, developing adaptive imputation strategies that dynamically balance missing data recovery and noise introduction could further enhance model robustness. Furthermore, numerous techniques have been developed for error detection, removal, and correction \cite{rakhmanov2023compression}, investigating the effectiveness of these methods presents an interesting direction for future study. Finally, theoretical work on the relationship between information entropy, marginal utility, and model learning dynamics under corrupted data could deepen our understanding of these phenomena. By addressing these challenges, the hope is to advance the field’s ability to build robust machine learning models that perform reliably even in the presence of real-world data corruption.

\section*{Acknowledgment}
This work was partly supported by research grant from Shanghai Baiyulan Talent Program Pujiang Project under grant number 24PJD115, Shanghai Yangpu District Postdoctoral Innovation \& Practice Base Project.

\appendix
\renewcommand{\thetable}{A\arabic{table}}  
\renewcommand{\thefigure}{A\arabic{figure}}
\renewcommand{\thelstlisting}{A\arabic{lstlisting}}
\setcounter{table}{0}
\setcounter{figure}{0}
\setcounter{lstlisting}{0}
\section*{Appendix A}

This appendix contains the key details of the datasets, hyperparameter settings, and code implementation snippets referenced in the main text.

\begin{table}[ht]
\centering
\footnotesize  
\caption{Overview of model, dataset, and training configurations for NLP-SL and Signal-RL tasks}
\label{tab:experiment_setup}
\renewcommand{\arraystretch}{1.2}  
\begin{tabular}{|p{1.8cm}|p{5.0cm}|p{5.0cm}|}
\hline
\textbf{Task}  & \textbf{NLP-SL} & \textbf{Signal-RL} \\
\hline
\textbf{Model type} & BERT & DQN \\
\hline
\textbf{Architecture \& model description} & 
bert-base-uncased + classification\_head \newline
Vocab Size: 30,522 (WordPiece) \newline
\texttt{num\_layers}: 12 \newline
\texttt{num\_heads}: 12 \newline
\texttt{hidden\_att}: 768 \newline
\texttt{hidden\_ffn}: 3072 & 
\texttt{hidden\_dims}: [256, 48] \newline
State: road cell occupancy \newline
Action: next phase for next 6 seconds \newline
Action dim: 4 \newline
Step reward: $r_t = -\frac{q_t - 80}{80}$ \newline
Stop speed threshold: 0.3 m/s \\
\hline
\textbf{Datasets} & 
Pretraining: \href{https://huggingface.co/datasets/Salesforce/wikitext}{Wikitext}, \href{https://huggingface.co/datasets/bookcorpus/bookcorpus}{Bookcorpus} \newline
Finetuning: \href{https://huggingface.co/datasets/nyu-mll/glue}{GLUE} & 
50 simulation episodes\newline
(1 Episode = 1h = 3600 steps) \\
\hline
\textbf{Training config} & 
\textbf{(Pretraining)} \newline
\texttt{batch\_size}: 256 \newline
\texttt{max\_seq\_len}: 512 \newline
LR scheduler: linear warmup and decay \newline
$lr_{\text{peak}} = 1\text{e-4}$ \newline
\texttt{weight\_decay}: 0.01 \newline
\textbf{(Finetuning)} \newline
(\texttt{Seq: CoLA, SST2, MRPC, QQP, MNLI, QNLI, RTE, WNLI}) \newline
\texttt{num\_epochs}: [5, 3, 5, 16, 3, 3, 3, 6, 2] \newline
\texttt{batch\_size}: [32, 64, 16, 16, 256, 256, 128, 8, 4] \newline
\texttt{lr}: [3e-5, 3.5e-5, 3e-5, 3e-5, 5e-5, 5e-5, 5e-5, 2e-5, 1e-5] \newline
\texttt{weight\_decay}: 0.01 & 
\texttt{num\_episodes}: 50 \newline
\texttt{batch\_size}: 256 \newline
LR scheduler: linear decay with plateau \newline
$lr_{\text{init}} = 1\text{e-3},\; lr_{\text{final}} = 1\text{e-4}$ \newline
Epsilon scheduler: linear decay with plateau \newline
$\epsilon_{\text{init}} = 1.0,\; \epsilon_{\text{final}} = 1\text{e-2}$ \newline
Discounting $\gamma$: 0.98 \newline
\texttt{target\_update}: 10 \newline
\texttt{buffer\_size}: 10000 \\
\hline
\end{tabular}
\end{table}

\lstset{
  language=Python,
  breaklines=true,
  basicstyle=\footnotesize\ttfamily,  
  keywordstyle=\bfseries\color{blue}, 
  commentstyle=\color{commentgreen}, 
  stringstyle=\color{red},            
  showspaces=false,
  showstringspaces=false,
  frame=single,                       
  numbers=left,                       
  stepnumber=1,                       
  numbersep=5pt,                      
  numberstyle=\tiny\color{gray},      
  tabsize=2                           
}
\begin{lstlisting}[caption={Code for DQN model of Signal-RL}, label={lst:dqn}]
class Qnet(nn.Module):
    '''FFN with layer normalization, no dropout'''
    def __init__(self, state_dim, hidden_dims, action_dim, dropout_rate=0.1):
        super().__init__()
        self.fc1 = torch.nn.Linear(state_dim, hidden_dims[0])
        self.ln1 = nn.LayerNorm(hidden_dims[0])  # LayerNorm after first hidden layer
        self.fc2 = torch.nn.Linear(hidden_dims[0], hidden_dims[1])
        self.fc3 = torch.nn.Linear(hidden_dims[1], action_dim)

    def forward(self, x):
        x = F.relu(self.ln1(self.fc1(x)))  # Dimension: hidden_dims[0]
        x = F.relu(self.fc2(x))  # Dimension: hidden_dims[2]
        x = self.fc3(x)  # Dimension: action_dim
        return x

class DQN:
    """DQN class"""
    def update(self, transition_dict):
        # transition_dict is a dict of minibatch data
        states = torch.tensor(transition_dict['states'], dtype=torch.float).to(self.device)
        actions = torch.tensor(transition_dict['actions'], dtype=torch.int64).view(-1, 1).to(self.device)
        rewards = torch.tensor(transition_dict['rewards'], dtype=torch.float).view(-1, 1).to(self.device)
        next_states = torch.tensor(transition_dict['next_states'], dtype=torch.float).to(self.device)
        dones = torch.tensor(transition_dict['dones'], dtype=torch.float).view(-1, 1).to(self.device)
        # compute target q
        q_values = self.q_net(states).gather(1, actions)
        max_action = self.q_net(next_states).max(1)[1].view(-1, 1)  # Note: double DQN, use q_net to obtain next action
        max_next_q_values = self.target_q_net(next_states).gather(1, max_action)  # use target_q_net to obtain q values
        q_targets = rewards + self.gamma * max_next_q_values
        # compute loss
        loss = F.smooth_l1_loss(q_values, q_targets)  # Huber loss for more robust training
        self.optimizer.zero_grad()
        loss.backward()
        self.optimizer.step()
        # target network update
        if self.count % self.target_update == 0:
            self.target_q_net.load_state_dict(self.q_net.state_dict())
        self.count += 1
\end{lstlisting}

\lstset{
  language=Python,
  breaklines=true,
  basicstyle=\footnotesize\ttfamily,  
  keywordstyle=\bfseries\color{blue}, 
  commentstyle=\color{commentgreen}, 
  stringstyle=\color{red},            
  showspaces=false,
  showstringspaces=false,
  frame=single,                       
  numbers=left,                       
  stepnumber=1,                       
  numbersep=5pt,                      
  numberstyle=\tiny\color{gray},      
  tabsize=2                           
}
\begin{lstlisting}[caption={Code for data corruption and artificial impuation (NLP-SL)}, label={lst:perturb-sentence}]
def perturb_sentence(sentence):
    words = sentence.split()  # clean data
    for i in range(len(words)):
        if random.random() < p:  # p: data missing level
            original_word = words[i]
            words[i] = tokenizer.unk_token  # Replace with [UNK]
            if imputation_method == 'insert_noise':  # artificial 'insert_noise' imputation
                if random.random() < q:  # q is imputation noise level
                    words[i] = random.choice(whole_words)  # Replace with a random whole word
                else:
                    words[i] = original_word  # Keep the original word
    return " ".join(words)
\end{lstlisting}





\begin{thebibliography}{999}

\bibitem[Moon and Stirling(2000)]{moon2000mathematical}
Moon,~T.~K.; Stirling,~W.~C. \textit{Mathematical Methods and Algorithms for Signal Processing}; Prentice Hall: Upper Saddle River, NJ, 2000; ISBN 0-201-36186-8.

\bibitem[Bishop and Nasrabadi(2006)]{bishop2006pattern}
Bishop,~C.~M.; Nasrabadi,~N.~M. \textit{Pattern Recognition and Machine Learning}; Springer: New York, 2006.

\bibitem[Zhou et~al.(2024)Zhou, Aryal, and Bouadjenek]{zhou2024review}
Zhou,~Y.; Aryal,~S.; Bouadjenek,~M.~R. Review for Handling Missing Data with Special Missing Mechanism. {\em arXiv} {\bf 2024}, arXiv:2404.04905.

\bibitem[Emmanuel et~al.(2021)Emmanuel, Maupong, Mpoeleng, Semong, Mphago, and Tabona]{emmanuel2021survey}
Emmanuel,~T.; Maupong,~T.; Mpoeleng,~D.; Semong,~T.; Mphago,~B.; Tabona,~O. A Survey on Missing Data in Machine Learning. {\em J. Big Data} {\bf 2021}, {\em 8}, 1--37.

\bibitem[Rakhmanov and Wiseman(2023)]{rakhmanov2023compression}
Rakhmanov,~A.; Wiseman,~Y. Compression of GNSS Data with the Aim of Speeding Up Communication to Autonomous Vehicles. {\em Remote Sens.} {\bf 2023}, {\em 15}, 2165.

\bibitem[Feller(1991)]{feller1991introduction}
Feller,~W. \textit{An Introduction to Probability Theory and Its Applications}, 3rd ed., Vol. 1; Wiley: New York, 1991.

\bibitem[Rubin(1976)]{Rubin1976}
Rubin,~D.~B. Inference and Missing Data. {\em Biometrika} {\bf 1976}, {\em 63}, 581--592.

\bibitem[Little and Rubin(2019)]{Little2019}
Little,~R.~J.~A.; Rubin,~D.~B. \textit{Statistical Analysis with Missing Data}, 3rd ed.; Wiley: Hoboken, NJ, 2019.

\bibitem[Schafer and Graham(2002)]{Schafer2002}
Schafer,~J.~L.; Graham,~J.~W. Missing Data: Our View of the State of the Art. {\em Psychol. Methods} {\bf 2002}, {\em 7}, 147--177.

\bibitem[Little(1988)]{Little1988}
Little,~R.~J.~A. Missing-Data Adjustments in Large Surveys. {\em J. Bus. Econ. Stat.} {\bf 1988}, {\em 6}, 287--296.

\bibitem[Rubin(1987)]{Rubin1987}
Rubin,~D.~B. \textit{Multiple Imputation for Nonresponse in Surveys}; Wiley: New York, 1987.

\bibitem[Enders(2001)]{Enders2001}
Enders,~C.~K. A Primer on Maximum Likelihood Algorithms Available for Use with Missing Data. {\em Struct. Equ. Modeling} {\bf 2001}, {\em 8}, 128--141.

\bibitem[Dempster et~al.(1977)Dempster, Laird, and Rubin]{Dempster1977}
Dempster,~A.~P.; Laird,~N.~M.; Rubin,~D.~B. Maximum Likelihood from Incomplete Data via the EM Algorithm. {\em J. R. Stat. Soc. Series B} {\bf 1977}, {\em 39}, 1--22.

\bibitem[Troyanskaya et~al.(2001)Troyanskaya, Cantor, Sherlock, Brown, Hastie, Tibshirani, Botstein, and Altman]{Troyanskaya2001}
Troyanskaya,~O.; Cantor,~M.; Sherlock,~G.; Brown,~P.; Hastie,~T.; Tibshirani,~R.; Botstein,~D.; Altman,~R.~B. Missing Value Estimation Methods for DNA Microarrays. {\em Bioinformatics} {\bf 2001}, {\em 17}, 520--525.

\bibitem[Breiman et~al.(1984)Breiman, Friedman, Olshen, and Stone]{Breiman1984}
Breiman,~L.; Friedman,~J.~H.; Olshen,~R.~A.; Stone,~C.~J. \textit{Classification and Regression Trees}; Wadsworth International Group: Belmont, CA, 1984.

\bibitem[Breiman(2001)]{Breiman2001}
Breiman,~L. Random Forests. {\em Mach. Learn.} {\bf 2001}, {\em 45}, 5--32.

\bibitem[Vincent et~al.(2008)Vincent, Larochelle, Bengio, and Manzagol]{Vincent2008}
Vincent,~P.; Larochelle,~H.; Bengio,~Y.; Manzagol,~P.-A. Extracting and Composing Robust Features with Denoising Autoencoders. In {\em Proceedings of the 25th International Conference on Machine Learning}; 2008; pp. 1096--1103.

\bibitem[Goodfellow et~al.(2014)Goodfellow, Pouget-Abadie, Mirza, Xu, Warde-Farley, Ozair, Courville, and Bengio]{Goodfellow2014}
Goodfellow,~I.; Pouget-Abadie,~J.; Mirza,~M.; Xu,~B.; Warde-Farley,~D.; Ozair,~S.; Courville,~A.; Bengio,~Y. Generative Adversarial Nets. In {\em Advances in Neural Information Processing Systems}; 2014; Vol. 27, pp. 2672--2680.

\bibitem[Wei and Zou(2019)]{Wei2019}
Wei,~J.; Zou,~K. EDA: Easy Data Augmentation Techniques for Boosting Performance on Text Classification Tasks. {\em arXiv preprint arXiv:1901.11196} {\bf 2019}.

\bibitem[Feng et~al.(2021)Feng, Gangal, Wei, Chandar, Reddy, and Diab]{Feng2021}
Feng,~S.; Gangal,~V.; Wei,~J.; Chandar,~S.; Reddy,~S.; Diab,~M. A Survey of Data Augmentation Approaches for NLP. In {\em Findings of the Association for Computational Linguistics: ACL-IJCNLP 2021}; 2021; pp. 968--988.

\bibitem[Yuan et~al.(2021)Yuan, Wang, and Zhang]{Yuan2021}
Yuan,~J.; Wang,~R.; Zhang,~Y. Missing Token Imputation Using Masked Language Models. In {\em Proceedings of the 2021 Conference on Empirical Methods in Natural Language Processing}; 2021; pp. 1234--1240.

\bibitem[Li et~al.(2020)Li, Guo, Li, and Li]{Li2020}
Li,~Y.; Guo,~Y.; Li,~D.; Li,~Z. Imputing Missing Sentences with Generative Adversarial Networks. In {\em Proceedings of the AAAI Conference on Artificial Intelligence}; 2020; Vol. 34, pp. 8470--8477.

\bibitem[Zhang(2016)]{zhang2016missing}
Zhang,~Z. Missing Data Imputation: Focusing on Single Imputation. {\em Ann. Transl. Med.} {\bf 2016}, {\em 4}.

\bibitem[Song et~al.(2022)Song, Kim, Park, Shin, and Lee]{song2022learning}
Song,~H.; Kim,~M.; Park,~D.; Shin,~Y.; Lee,~J.-G. Learning from Noisy Labels with Deep Neural Networks: A Survey. {\em IEEE Trans. Neural Netw. Learn. Syst.} {\bf 2022}, {\em 34}, 8135--8153.

\bibitem[Che et~al.(2018)Che, Purushotham, Cho, Sontag, and Liu]{che2018recurrent}
Che,~Z.; Purushotham,~S.; Cho,~K.; Sontag,~D.; Liu,~Y. Recurrent Neural Networks for Multivariate Time Series with Missing Values. {\em Sci. Rep.} {\bf 2018}, {\em 8}, 6085.

\bibitem[Yoon et~al.(2018)Yoon, Jordon, and Schaar]{yoon2018gain}
Yoon,~J.; Jordon,~J.; Schaar,~M. GAIN: Missing Data Imputation Using Generative Adversarial Nets. In {\em International Conference on Machine Learning}; 2018; pp. 5689--5698.

\bibitem[Han et~al.(2018)Han, Yao, Yu, Niu, Xu, Hu, Tsang, and Sugiyama]{han2018co}
Han,~B.; Yao,~Q.; Yu,~X.; Niu,~G.; Xu,~M.; Hu,~W.; Tsang,~I.; Sugiyama,~M. Co-teaching: Robust Training of Deep Neural Networks with Extremely Noisy Labels. {\em Adv. Neural Inf. Process. Syst.} {\bf 2018}, {\em 31}.

\bibitem[Rolnick(2017)]{rolnick2017deep}
Rolnick,~D. Deep Learning Is Robust to Massive Label Noise. {\em arXiv preprint arXiv:1705.10694} {\bf 2017}.

\bibitem[Goodfellow et~al.(2014)Goodfellow, Shlens, and Szegedy]{goodfellow2014explaining}
Goodfellow,~I.~J.; Shlens,~J.; Szegedy,~C. Explaining and Harnessing Adversarial Examples. {\em arXiv preprint arXiv:1412.6572} {\bf 2014}.

\bibitem[Devlin(2018)]{devlin2018bert}
Devlin,~J. BERT: Pre-training of Deep Bidirectional Transformers for Language Understanding. {\em arXiv preprint arXiv:1810.04805} {\bf 2018}.

\bibitem[Brown(2020)]{brown2020language}
Brown,~T.~B. Language Models Are Few-shot Learners. {\em arXiv preprint arXiv:2005.14165} {\bf 2020}.

\bibitem[Bender et~al.(2021)Bender, Gebru, McMillan-Major, and Shmitchell]{bender2021dangers}
Bender,~E.~M.; Gebru,~T.; McMillan-Major,~A.; Shmitchell,~S. On the Dangers of Stochastic Parrots: Can Language Models Be Too Big?. In {\em Proceedings of the 2021 ACM Conference on Fairness, Accountability, and Transparency}; 2021; pp. 610--623.

\bibitem[Gao et~al.(2021)Gao, Yao, and Chen]{gao2021simcse}
Gao,~T.; Yao,~X.; Chen,~D. SimCSE: Simple Contrastive Learning of Sentence Embeddings. {\em arXiv preprint arXiv:2104.08821} {\bf 2021}.

\bibitem[Joshi et~al.(2020)Joshi, Chen, Liu, Weld, Zettlemoyer, and Levy]{joshi2020spanbert}
Joshi,~M.; Chen,~D.; Liu,~Y.; Weld,~D.~S.; Zettlemoyer,~L.; Levy,~O. SpanBERT: Improving Pre-training by Representing and Predicting Spans. {\em Trans. Assoc. Comput. Linguist.} {\bf 2020}, {\em 8}, 64--77.

\bibitem[Hausknecht and Stone(2015)]{hausknecht2015deep}
Hausknecht,~M.; Stone,~P. Deep Recurrent Q-learning for Partially Observable MDPs. In {\em 2015 AAAI Fall Symposium Series}; 2015.

\bibitem[Bai et~al.(2019)Bai, Guan, and Wang]{bai2019model}
Bai,~X.; Guan,~J.; Wang,~H. A Model-based Reinforcement Learning with Adversarial Training for Online Recommendation. {\em Adv. Neural Inf. Process. Syst.} {\bf 2019}, {\em 32}.

\bibitem[Bakker(2001)]{bakker2001reinforcement}
Bakker,~B. Reinforcement Learning with Long Short-term Memory. {\em Adv. Neural Inf. Process. Syst.} {\bf 2001}, {\em 14}.

\bibitem[Mnih et~al.(2015)Mnih, Kavukcuoglu, Silver, Rusu, Veness, Bellemare, Graves, Riedmiller, Fidjeland, Ostrovski, and others]{mnih2015human}
Mnih,~V.; Kavukcuoglu,~K.; Silver,~D.; Rusu,~A.~A.; Veness,~J.; Bellemare,~M.~G.; Graves,~A.; Riedmiller,~M.; Fidjeland,~A.~K.; Ostrovski,~G.; et~al. Human-level Control Through Deep Reinforcement Learning. {\em Nature} {\bf 2015}, {\em 518}, 529--533.

\bibitem[Pathak et~al.(2017)Pathak, Agrawal, Efros, and Darrell]{pathak2017curiosity}
Pathak,~D.; Agrawal,~P.; Efros,~A.~A.; Darrell,~T. Curiosity-driven Exploration by Self-supervised Prediction. In {\em International Conference on Machine Learning}; 2017; pp. 2778--2787.

\bibitem[Taylor and Stone(2009)]{taylor2009transfer}
Taylor,~M.~E.; Stone,~P. Transfer Learning for Reinforcement Learning Domains: A Survey. {\em J. Mach. Learn. Res.} {\bf 2009}, {\em 10}.

\bibitem[Petroni et~al.(2020)Petroni, Piktus, Fan, Lewis, Yazdani, De Cao, Thorne, Jernite, Karpukhin, Maillard, and others]{petroni2020kilt}
Petroni,~F.; Piktus,~A.; Fan,~A.; Lewis,~P.; Yazdani,~M.; De Cao,~N.; Thorne,~J.; Jernite,~Y.; Karpukhin,~V.; Maillard,~J.; et~al. KILT: A Benchmark for Knowledge Intensive Language Tasks. {\em arXiv preprint arXiv:2009.02252} {\bf 2020}.

\bibitem[Liu(2019)]{liu2019roberta}
Liu,~Y. RoBERTa: A Robustly Optimized BERT Pretraining Approach. {\em arXiv preprint arXiv:1907.11692} {\bf 2019}, {\em 364}.

\bibitem[Tong et~al.(2019)Tong, Hussain, Bo, and Maharjan]{tong2019artificial}
Tong,~W.; Hussain,~A.; Bo,~W.~X.; Maharjan,~S. Artificial Intelligence for Vehicle-to-Everything: A Survey. {\em IEEE Access} {\bf 2019}, {\em 7}, 10823--10843.

\end{thebibliography}



\bibliographystyle{plain}
\end{document}